\documentclass[
aip,
jcp,
reprint,
raggedbottom
]{revtex4-2}

\usepackage[T1]{fontenc}
\usepackage[utf8]{inputenc}
\usepackage{amsmath}
\usepackage{amssymb}
\usepackage{graphicx}
\usepackage{xcolor}
\usepackage{siunitx}
\DeclareSIUnit\angstrom{\text{\AA}}
\definecolor{linkblue}{RGB}{80, 100, 255}
\usepackage{hyperref}
\hypersetup{
    colorlinks=true,
    linkcolor=linkblue,
    filecolor=linkblue,      
    urlcolor=linkblue,
    citecolor=linkblue,
}

\usepackage{booktabs}
\usepackage{pifont}
\usepackage{array}
\usepackage{tabularx}
\usepackage{capt-of}
\newcolumntype{W}{p{11.5cm}}
\usepackage{algpseudocode}

\makeatletter
\newcounter{algorithm}
\renewcommand{\thealgorithm}{\arabic{algorithm}}
\newenvironment{algorithm}[1][t]
  {\begin{figure}[#1]\refstepcounter{algorithm}%
   \noindent\rule{\linewidth}{0.9pt}\vspace{1pt}\par\footnotesize}
  {\par\vspace{-2pt}\noindent\rule{\linewidth}{0.9pt}\end{figure}}
\newcommand{\algcaption}[1]{%
  \par\vspace{1pt}\noindent\rule{\linewidth}{0.4pt}\par\vspace{2pt}%
  \noindent\textbf{ALG.~\thealgorithm.}~#1\par\vspace{2pt}}
\makeatother

\DeclareMathOperator{\E}{\mathbf{E}}

\DeclareMathOperator*{\argmax}{arg\,max}

\newcommand{\fitcol}[1]{%
  \resizebox{\ifdim\width>\linewidth\linewidth\else\width\fi}{!}{#1}}

\newcommand{\methodname}{\textsf{ARCHER}}
\newcommand{\Ang}{\text{\AA}}
\newcommand{\SO}{\mathrm{SO}(3)}
\newcommand{\CTF}{\mathcal{H}}
\newcommand{\Slice}{\mathcal{S}}
\newcommand{\Rot}{\mathbf{R}}
\newcommand{\Vol}{V}
\newcommand{\geo}{d_{\mathrm{geo}}}

\newcommand{\maintitle}{\methodname{}: Amortized cross-specimen pose estimation
for cryo-electron microscopy}

\begin{document}

\title{\maintitle}

\author{Nhan D. Nguyen}
\email{ndnguyen@uchicago.edu}
\affiliation{Pritzker School of Molecular Engineering, University of Chicago, Chicago, Illinois 60637, United States}
\author{Bao Pham}
\affiliation{Department of Computer Science, Rensselaer Polytechnic Institute, Troy, New York 12180, United States}

\begin{abstract}
Single-particle cryo-electronic microscopy (cryo-EM) pose estimation is traditionally solved anew for each dataset,  where iterative refinement is done from scratch while the estimator learns to store the molecule in its weights. In this work, we show that pose inference is a generalizable, specimen-agnostic operation when conditioned explicitly on a reference volume. We introduce \textbf{\methodname{}}, an amortized contrastive classifier that models the pose posterior over a discrete rotation grid. Trained across a variety of protein structures, it operates zero-shot \textit{without retraining per structure}. This transferability is grounded in Fourier-space information mechanics, where all specimen dependence is captured by the reference structure's power spectrum and spatial extent. \methodname{} achieves a median angular error of $5.0^\circ$ on $100$ held-out test structures and $2.5^\circ$ on experimental particles, matching dedicated estimators within $0.16\text{ \AA}$ in 3D reconstruction. Crucially, downstream conformational signal is preserved. The leading conformational coordinate correlates at $0.97$ with deposited benchmarks, faithfully reconstructing free-energy basins and mobile domains. These results overall demonstrate that cryo-EM pose estimation can be generalized across different structures.
\end{abstract}
\maketitle


\section{Introduction}\label{sec:intro}

\begin{figure*}[!ht]
\centering
\includegraphics[width=\textwidth,height=1\textheight,keepaspectratio]{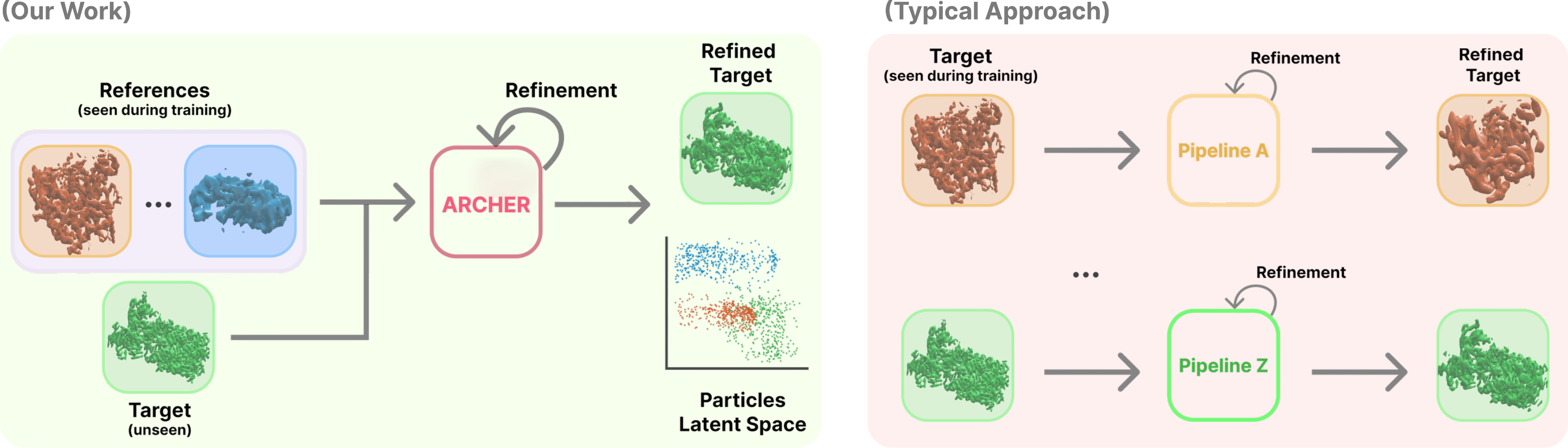}
\caption{\textbf{Comparison between \methodname{} and typical approaches.} \methodname{} takes the reference volume as an \emph{argument}. It is trained once across many structures and then applied to a target it has never seen, whose reference is handed to it at inference time alongside the particles. Refinement is applied afterwards on the estimate it produced, where the particle embeddings it computed support downstream heterogeneity analysis. In other methods, the specimen is absorbed into the weights, so each target requires its own fitted model and its own training run, and nothing learned on one target is transferred to the next.}
\label{fig:contrast}
\end{figure*}

\begin{figure*}[!ht]
\centering
\includegraphics[width=\textwidth,height=0.75\textheight,keepaspectratio]{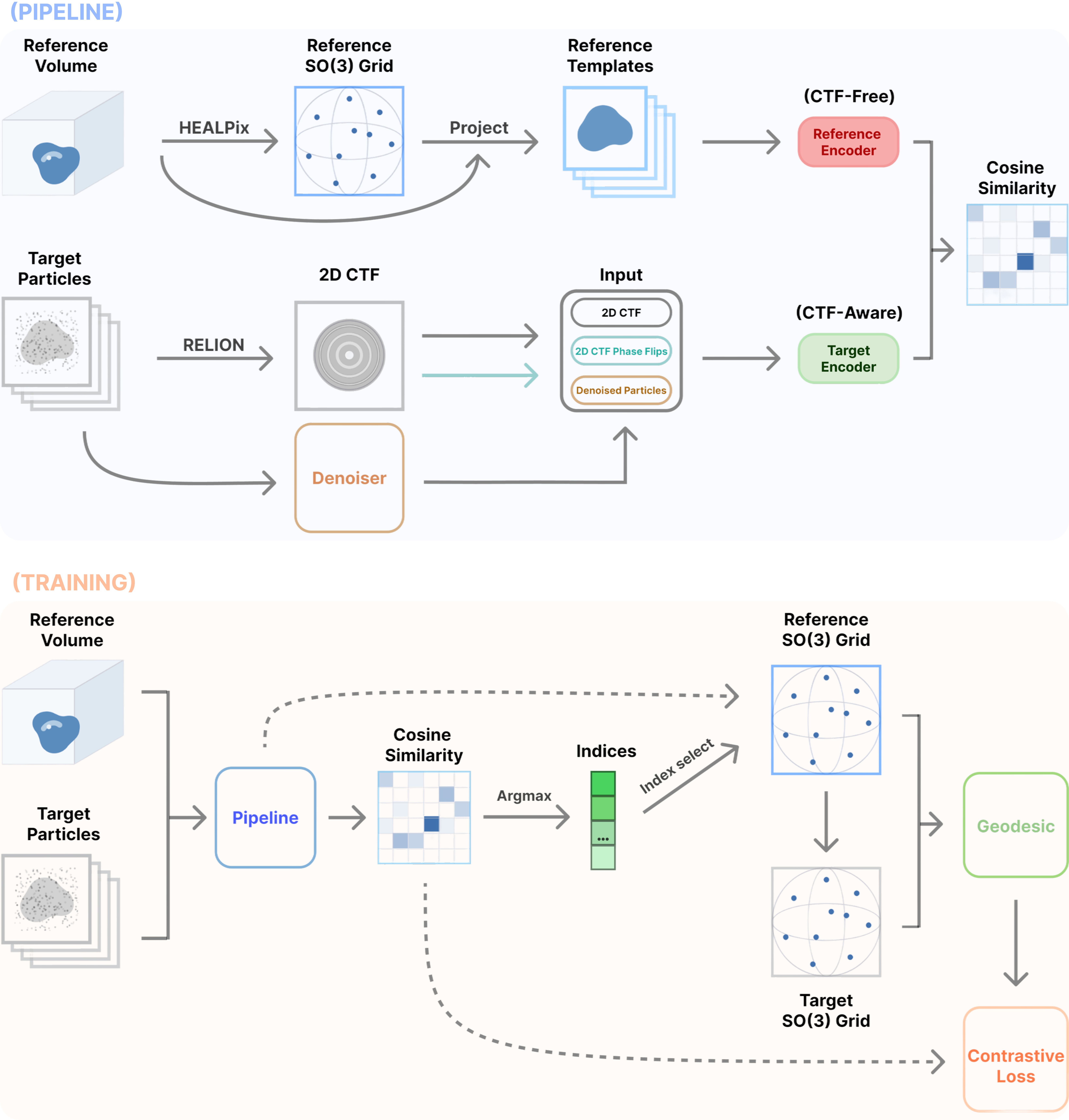}
\caption{\textbf{Overview of ARCHER's architecture.}~\emph{Pipeline}: The reference volume is sliced at every cell of a HEALPix grid on $\SO$, giving a fixed bank of templates that is embedded \emph{once} per specimen. Each particle is embedded by a second encoder from a stack of image channels: the recorded particle, its phase-flipped copy, and denoised particles. The transfer function enters the network through the pixels it modulates. The pose $\Rot$ is the grid cell whose template embedding maximizes a temperature-scaled cosine similarity against the particle embedding.
\emph{Training}: The same forward pass is run on simulated particles
whose true rotation is known. That rotation is mapped to its nearest grid cell,
and the resulting cosine similarities are trained against a target smoothed over
the geodesic neighborhood of that cell. The two encoders share a topology and hold independent weights.}
\label{fig:arch}
\end{figure*}

Single-particle cryo-electron microscopy (cryo-EM) determines macromolecular structure from images of individual molecules frozen in vitreous ice \cite{frank2006three,derosier1968reconstruction}. In recent years, direct electron detectors and improved reconstruction algorithms brought the method to atomic resolution across a wide range of specimens \cite{kuhlbrandt2014resolution,nakane2020single,yip2020atomic},
and it is now the principal technique for assemblies that resist
crystallization. However, the measurement is generally destructive and the electron dose is severely limited, so each particle is recorded at a signal-to-noise ratio well below unity and structure is recovered by averaging $10^5$ to $10^6$ of the raw particles.

Averaging requires knowing how each particle is oriented. Because the specimen is frozen in random orientations, the rotation that produced each image is a latent variable that must be inferred jointly with the structure\cite{sigworth1998maximum,scheres2012bayesian,bendory2020cryoem}. By the central-slice theorem \cite{crowther1970reconstruction}, a projection samples the Fourier transform of the volume on a plane through the origin, where recovering the volume amounts to placing every measured plane correctly in three dimensions. Errors in that placement blur the average, and the placement
problem dominates both the computational cost and the failure modes of the pipeline \cite{scheres2012relion,punjani2017cryosparc}.

\textbf{Iterative refinement.}~The established solution treats poses as missing data and marginalizes them. Maximum-likelihood formulations \cite{sigworth1998maximum,penczek1994ribosome} and their Bayesian extension \cite{scheres2012bayesian} alternate between computing a posterior over orientations for every particle and rebuilding the volume as a posterior-weighted backprojection, an expectation--maximization scheme \cite{dempster1977maximum} implemented in RELION
\cite{scheres2012relion,kimanius2021relion4}, cryoSPARC \cite{punjani2017cryosparc} and FREALIGN \cite{grigorieff2007frealign}. Reliability comes from evaluating half-sets independently and reporting a Fourier shell correlation between them \cite{scheres2012goldstandard,henderson2012outcome}, with resolution read at a fixed criterion \cite{rosenthal2003optimal} or one that accounts for the number of samples per shell \cite{vanheel2005fourier}. This machinery is mature, and it restarts from scratch for every dataset: the scoring operation is recomputed from first principles each time, at a cost that scales with the product of particles and candidate orientations.

\textbf{Learned estimators.}~ Neural approaches replace part of that loop. One family performs amortized inference within a single dataset of a single structure: cryoDRGN \cite{zhong2021cryodrgn,zhong2021cryodrgn2} learns a conformational latent space with a coordinate-based decoder, while cryoAI and cryoFIRE \cite{levy2022cryoai,levy2022cryofire} add an encoder that predicts pose directly, removing the per-particle search. cryoSPIN \cite{shekarforoush2024cryospin} refines an initial prediction by search, and other related works explore implicit \cite{murphy2021implicit,qu2025cryonerf} and diffusion-based \cite{li2024cryoddm} representations. A second family is
supervised: given a solved structure and its assigned poses, a rotation classifier can be trained and applied to the same specimen, as in cryoPARES \cite{sanchezgarcia2025cryopares} and earlier work on learned orientation assignment \cite{banjac2021learning,lian2022end}. The CESPED benchmark \cite{sanchezgarcia2023cesped} standardized this setting. A third line targets ab-initio determination without any reference, either by amortized regression as in CryoFastAR \cite{zhang2025cryofastar} or by classical common-lines synchronization \cite{singer2011three,wang2013orientation,wang2013firm}.

Across these families of approaches, the volume lives in the network parameters. An encoder
trained on one specimen encodes that specimen's projections, so a new protein
requires new training, and the computation spent learning to compare a noisy
image against a candidate view is discarded. Recent foundation-model work in
cryo-EM has begun to target reusable components rather than per-dataset
models: cryoFM \cite{zhou2025cryofm} learns a generative prior over densities,
Cryo-IEF \cite{yan2026cryoief} learns particle features by self-supervision,
and learned regularization \cite{kimanius2024blush} and map restoration
\cite{he2023emready,sanchezgarcia2021deepemhancer} transfer across specimens.
Pose assignment itself has remained per-dataset. In this work, we instead ask whether the comparison at the heart of pose assignment can be learned once and reused. 

The quantity that every pose estimator evaluates is the posterior
$p(\Rot \mid y, \Vol)$ over rotations, in which $y\in\mathbb{R}^{L\times L}$ is a single recorded particle image and $\Vol\in\mathbb{R}^{L\times L\times L}$ is the reference volume on a cubic grid of side $L$ voxels. When $\Vol$ is supplied as an argument rather than fitted into the weights, this expression separates two quantities that per-dataset methods entangle. The first is the specimen, which is data and differs at every target. The second is the operation of matching a noisy projection to a candidate view, which is governed by the microscope -- the contrast transfer function (CTF) \cite{rohou2015ctffind4}, the noise spectrum, and the slice geometry -- and is therefore common to every specimen. Provided with the reference volume, the network we develop learns a matching function that transfers to molecules it has never seen, and such a reference volume is almost always available in practice, whether as a consensus map, a homologue, or a predicted structure
\cite{jumper2021alphafold,lin2023esmfold,proteinrediff,liu2023kinase}.

\textbf{\methodname{}}. Our approach, \textbf{\emph{A}}mortized
\textbf{\emph{R}}eference-\textbf{\emph{c}}onditioned
\textbf{\emph{H}}ierarchical \textbf{\emph{R}}efinement realizes this posterior
as a contrastive classifier over a discrete rotation grid, and we train it
across $3{,}330$ protein structures and apply it without retraining. It places
particles at a median error of $5.0^\circ$ on $100$ structures held out of
training, and at $2.5^\circ$ on experimental data from EMPIAR-10076
\cite{iudin2016empiar}. Its reconstructions come within $0.16$~\AA{} of
per-dataset estimators where a target's own sampling is the binding constraint.
Conformational signal is left intact, in that the leading conformational
coordinate recovered through our poses agrees with the one recovered through the
deposited poses at a canonical correlation of $0.97$.

We build a diffusion forward process that interpolates the identity into the CTF, so that a recorded particle is \emph{exactly} its terminal state and denoising is one network evaluation at a known noise level rather than a sampled reverse trajectory -- which is cheap enough to sit inside the training loop and differentiable through it. Its measured effect on pose accuracy is confined to fine precision on held-out synthetic proteins (see Sec. (~\ref{sec:methods:diffusion})). Every result reported here is from the three-channel configuration.

Lastly, we quantified the amortization capability of our network. Rotating a structure
by a small angle displaces a Fourier component at radius $k$ by an arc
proportional to $k$, so the information about orientation grows as $k^2$ times
the spectral signal-to-noise ratio. This scaling follows from the microscope and
from the geometry of the central-slice theorem, and the specimen enters it
through its own power spectrum, which the reference volume supplies. That shared
physics is what allows one encoder to learn from many maps, and it sets both the
accuracy attainable and the point beyond which different methods converge on the
same range of accuracy.

\section{Related work}\label{sec:background}

\begin{figure*}[!t]
\centering
\includegraphics[width=\textwidth,height=0.82\textheight,keepaspectratio]{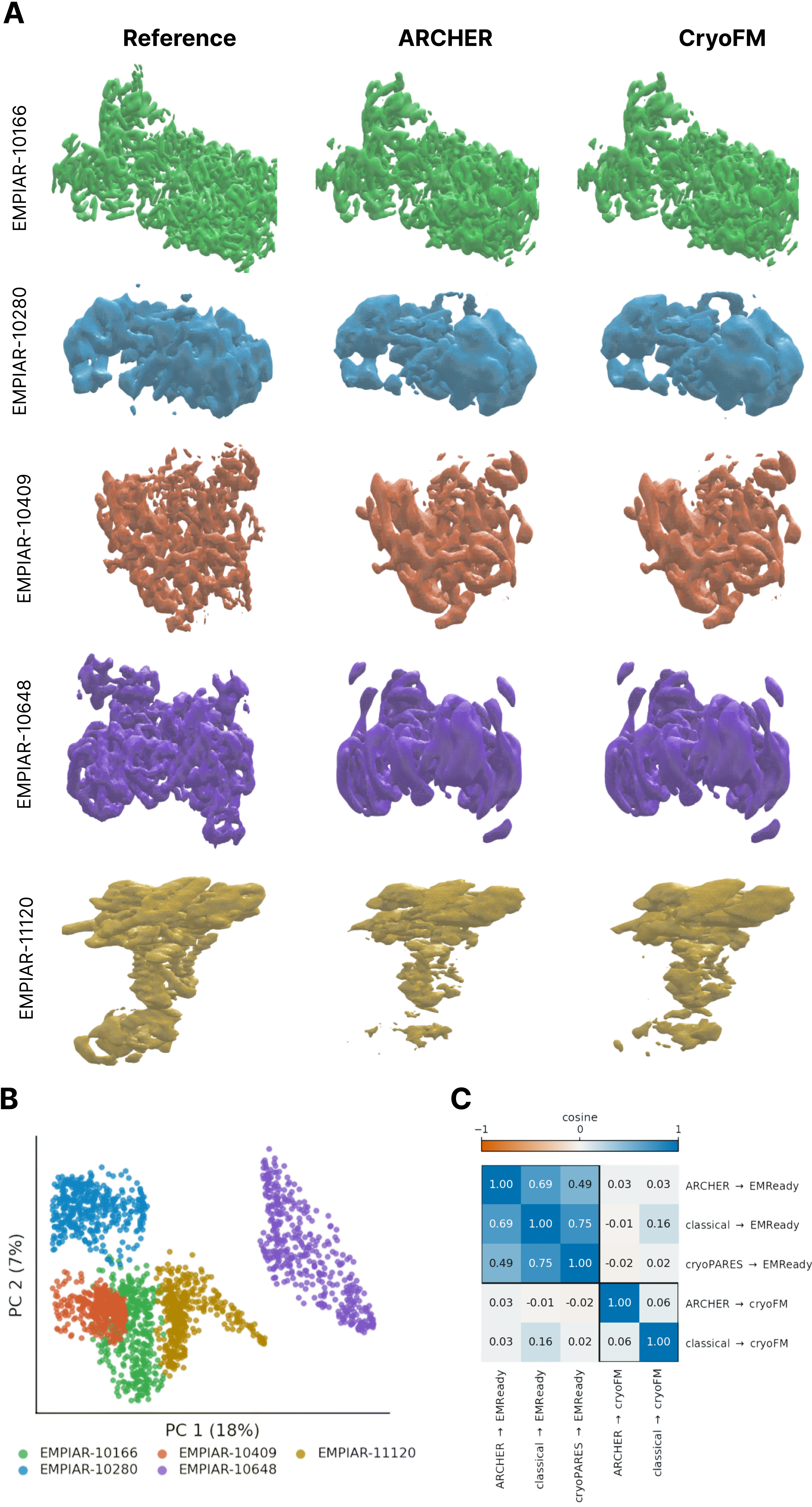}
\caption{\textbf{Reconstructions and restorations across specimen.} (A)~Isosurfaces for four CESPED targets, one row per target
and one column per map: the deposited reference, the \methodname{}
reconstruction, and that reconstruction after cryoFM restoration
\cite{zhou2025cryofm}. Color identifies the specimen; levels enclose a fixed
fraction of the mask volume within each row, that is, among maps of the same
molecule.
(B)~Every point is one experimental particle image embedded by principal
component analysis, colored by dataset. The four specimens occupy
distinguishable regions, and the separation is structural rather than
instrumental: repeating the embedding on phase-scrambled particles, which
preserves each image's power spectrum while destroying its structure, collapses
the between-specimen separation from $1.06$ to $0.05$ in units of the
within-specimen spread.
(C)~Cosines between restoration displacement vectors in full voxel space. A
restorer moves every map it is given in nearly the same direction, whichever
estimator produced that map, so restoration acts on the map rather than on the
pose estimate that made it. The same reconstructions embedded by method are
Fig.~\ref{fig:smethod}.}
\label{fig:maps}
\end{figure*}

\subsection{Orientation as a Latent Variable}\label{sec:bg:classical}

The reconstruction problem was posed in its modern form by De Rosier and Klug
\cite{derosier1968reconstruction} and Crowther \emph{et al.}
\cite{crowther1970reconstruction}: each micrograph is a line integral through
the specimen, so its Fourier transform is a central section of the transform of
the volume, and the structure follows once enough sections are placed
correctly. With orientations unknown, placement and reconstruction are coupled.

Two classical strategies resolve the coupling. Projection matching scores every particle against reference projections of a working volume and iterates \cite{penczek1994ribosome,grigorieff2007frealign}, performing an exhaustive search with a whitened matched filter and an explicit false-positive threshold, which detects and orients individual molecules in crowded images \cite{rickgauer2017single,lucas2021locating}. This is the classical pipeline we compare against throughout. 

Meanwhile, maximum-likelihood formulations instead treat the pose as a latent variable and marginalize it \cite{sigworth1998maximum}, which Scheres \cite{scheres2012bayesian} casts in a Bayesian framework with a regularizing prior on the volume and implemented as
expectation-maximization \cite{dempster1977maximum} in RELION
\cite{scheres2012relion,kimanius2021relion4}; cryoSPARC added stochastic gradient descent for initialization and branch-and-bound search \cite{punjani2017cryosparc}, later with adaptive regularization \cite{punjani2020nonuniform}. A third route avoids a reference entirely by exploiting the common-line geometry of pairs of projections and solving the resulting synchronization problem \cite{singer2011three,wang2013orientation,wang2013firm}.

Overall, these methods share a computational signature. The pose posterior is recomputed from first principles for every particle of every dataset, at a cost proportional to the number of candidate orientations, and nothing learned on one specimen is transferrable to another specimen. 

\subsection{Conformational Heterogeneity}\label{sec:bg:hetero}

Most interesting specimens are flexible, and the pose problem is entangled with a conformational one. Discrete treatments assign particles to a small number of classes \cite{scheres2010classification,lyumkis2013likelihood}, while multi-body refinement partitions the molecule into rigid units with independent orientations \cite{nakane2018multibody}. Continuous treatments learn a low-dimensional latent space: cryoDRGN couples an encoder to a coordinate-based volume decoder
\cite{zhong2021cryodrgn,zhong2021cryodrgn2}, 3DFlex models deformation fields directly \cite{punjani2023threedflex}, and RECOVAR\cite{gilles2025recovar} estimates the covariance of
the volume distribution and deconvolves the per-particle posterior to recover a conformational density. The last property matters for
evaluation: because RECOVAR returns an embedding with per-particle uncertainty and is deterministic given its inputs, running it twice on identical particles under different pose estimates isolates the effect of the poses, which is the comparison we use in Sec. (~\ref{sec:res:hetero}).

Heterogeneity also sets the difficulty of pose assignment itself. A flexible molecule presents projections that no single rigid volume explains, so scoring against a consensus map is systematically mismatched -- the regime in which estimators differ most.

\subsection{Learned Pose Estimation}\label{sec:bg:learned}

\emph{Per-structure Encoders.} Following the amortized-inference pattern of variational autoencoders (VAEs) \cite{kingma2014vae}, cryoAI and cryoFIRE replace the per-particle search with a VAE trained jointly with a volume representation \cite{levy2022cryoai,levy2022cryofire}, and cryoSPIN adds a search-based correction to the encoder's prediction
\cite{shekarforoush2024cryospin}. These methods amortize over the particles of a single dataset pertaining to a structure, so the training cost is repaid across images of the same specimen. Replacing search with a forward pass trades exactness for speed, and the resulting amortization gap between the inferred posterior and the true one is a known cost of the design \cite{cremer2018amortization}. Because the volume lives in
the decoder, the encoder is meaningful only for the targeted dataset it was fit on.

\emph{Supervised, Structure-specific.} Given a solved structure and its assigned poses, orientation assignment becomes supervised learning on $\SO$ \cite{banjac2021learning,lian2022end}, standardized by the CESPED benchmark \cite{sanchezgarcia2023cesped} and carried to production accuracy by cryoPARES \cite{sanchezgarcia2025cryopares}. This family learns the same matching operator we do, but ties it to one molecule, so its cost recurs per specimen: training it on four CESPED targets took $90.5$ single-GPU hours here, against a single training run that serves all of them (see Tab.~\ref{tab:cparestime} in the Appendix).

A protocol difference is worth naming before any of these are compared. The cryoDRGN workflow, and its ab-initio successor\cite{levy2025cryodrgnai}, interpose an interactive curation step in which particles judged to be junk are filtered out before or between reconstruction rounds \cite{zhong2021cryodrgn,zhong2021cryodrgn2}. In contrast, our approach is an end-to-end process: every particle is scored, none is removed, and no manual intervention enters between the particles and the poses. Comparisons that hold the particle set fixed, as ours do, are therefore stricter on us than on a curated pipeline.

\emph{Reference-free Amortization.} CryoFastAR regresses poses directly from particle sets without a reference \cite{zhang2025cryofastar}, and related work explores implicit representations \cite{qu2025cryonerf} and diffusion models \cite{li2024cryoddm,huang2026cryonetrefine} for ab-initio determination. cryoDRGN-AI is the current state of the art in this setting, recovering both structure and motion without a starting model \cite{levy2025cryodrgnai}, and CryoBench supplies the standardized heterogeneity datasets on which such methods are now compared \cite{jeon2024cryobench}. This is the hardest setting and the one where heterogeneity costs most (see our comparisons in Sec.~\ref{sec:results}).

\emph{Representing distributions on $\SO$.} All learned estimators must place a distribution on a curved space. Equivariant architectures \cite{cohen2018spherical,esteves2018learning} and the Image2Sphere\cite{klee2023image2sphere} construction build the symmetry into the network, while implicit-PDF\cite{murphy2021implicit} represents the density by evaluating an energy at sampled rotations. We take the discretization route, with a HEALPix grid \cite{gorski2005healpix,yershova2010generating} that gives near uniform coverage and a controllable spacing, and recover sub-cell precision by a continuous refinement step rather than by a finer grid.

\subsection{Reusable components}\label{sec:bg:foundation}

Parts of the pipeline have already been shown to transfer across specimens. Particle picking \cite{bepler2019topaz} and micrograph preprocessing \cite{tegunov2019warp} are routinely handled by models trained once and applied broadly. On the map side, learned regularization inside refinement \cite{kimanius2024blush} and post-hoc restoration \cite{he2023emready,sanchezgarcia2021deepemhancer} generalize to structures never seen in training, and cryoFM learns a generative prior over densities usable as a plug-in for downstream tasks \cite{zhou2025cryofm}. That prior is defined over maps rather than over orientations, so cryoFM contributes to a pipeline at the reconstruction and restoration stages and produces no per-particle pose; it therefore appears here as a map restorer (Sec.~\ref{sec:res:maps}) and cannot be placed on the pose  accuracy axis at all, which is a difference in what the two models represent rather than a comparison either way. Cryo-IEF learns particle-level features by self-supervision across many datasets \cite{yan2026cryoief}, demonstrating that image statistics transfer even when structures do not.

Orientation assignment has stayed outside this trend, and the reason is structural rather than incidental: the quantity a pose estimator must know is the specimen, and every design so far has supplied it by fitting. Providing it as an argument is what makes the remaining operation -- the comparison of a noisy projection against a candidate view under a known transfer function -- shared across specimens, and therefore learnable once. Secs.~(\ref{sec:methods}) and (\ref{sec:results}) make that statement quantitative and show what it implies for the accuracy attainable.

\section{Methods}\label{sec:methods}

\begin{figure}[!h]
\centering
\includegraphics[width=\linewidth,height=0.30\textheight,keepaspectratio]{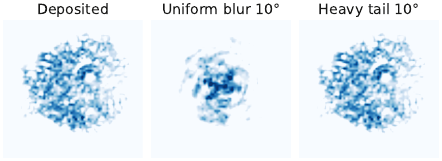}
\caption{\textbf{Error shape governs the map.} Reconstructions at matched mean
angular error under uniform and heavy-tailed pose perturbation.}
\label{fig:shape}
\end{figure}

\subsection{Forward model}\label{sec:methods:forward}
Volumes are sampled on a cubic grid of side $L$ voxels and images on the
corresponding $L\times L$ grid, following the discretization used in the
mathematical cryo-EM literature \cite{bendory2020cryoem,singer2020computational};
$L$ is reserved for this box side throughout and $C$ for the number of input
channels of a network. Table~\ref{tab:notation} in the Supplementary Material
displays every symbol with the corresponding space.

Let $\Vol\in\mathbb{R}^{L\times L\times L}$ be a volume,
$\hat{\Vol}=\mathcal{F}\Vol\in\mathbb{C}^{L\times L\times L}$ its discrete
Fourier transform, $\Rot\in\SO$ a rotation, and
$\xi\in\mathbb{R}^{3}$ the per-particle contrast-transfer parameters
(defocus along the two astigmatic axes and the astigmatism azimuth; voltage,
spherical aberration and amplitude contrast are fixed per dataset). A particle
image $y\in\mathbb{R}^{L\times L}$ is modeled as
\begin{equation}
y \;=\; \mathcal{F}^{-1}\!\bigl[\CTF_{\xi}\cdot \Slice_{\Rot}\hat{\Vol}\bigr]\;+\;\eta ,
\qquad
\Slice_{\Rot}\hat{\Vol}(\mathbf{q}) = \hat{\Vol}(\Rot^{-1}\mathbf{q}),
\label{eq:forward}
\end{equation}
where $\Slice_{\Rot}:\mathbb{C}^{L\times L\times L}\to\mathbb{C}^{L\times L}$
is the central-slice operator, $\CTF_{\xi}\in\mathbb{R}^{L\times L}$ the
contrast transfer function (CTF), $\mathbf{q}\in\mathbb{R}^{3}$ a Fourier
coordinate restricted to the slice plane
$\Pi_{\Rot}=\{\mathbf{q}:\mathbf{q}\cdot\Rot\hat{z}=0\}$, and
$\eta\in\mathbb{R}^{L\times L}$ colored Gaussian noise whose power in the shell
at radius $k=\lVert\mathbf{q}\rVert$ is $\sigma^{2}(k)$, measured from real
micrographs. The transfer function is astigmatic, with the sign convention of
Ref.~\cite{rohou2015ctffind4}. Equation~\eqref{eq:forward} is the flat
Ewald-sphere approximation \cite{derosier2000correction,wolf2006ewald}; it is
exact to the resolutions considered here and it makes handedness an exact
degeneracy, since $\Vol$ and its mirror produce identical projection sets
\cite{garciacondado2022handedness}.

\begin{figure*}[t]
\centering
\includegraphics[width=\textwidth,height=0.24\textheight,keepaspectratio]{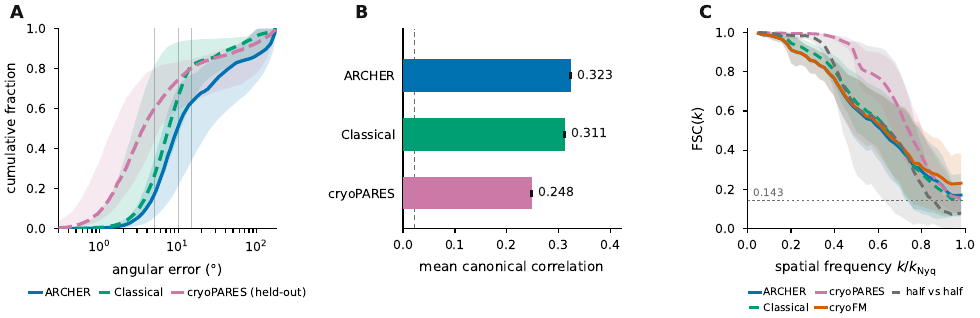}
\caption{\textbf{ARCHER performance on benchmark datasets used by per-structure estimators} (A)~Cumulative angular error over the five CESPED targets; lines
are the median across specimens, shading their full range. (B)~Retained
conformational signal on EMPIAR-10409, as the mean canonical correlation
over the four latent directions; the dashed line is the shuffled-pose floor.
(C)~Shell correlation pooled over the four targets of Fig.~\ref{fig:maps}A,
inside the benchmark mask; the abscissa is a fraction of each target's own
Nyquist frequency and the dashed line marks the $0.143$ criterion
\cite{rosenthal2003optimal}. Scoring conventions, matched particle counts and
error bars are given in App.~\ref{app:benchdetail}.}
\label{fig:bench}
\end{figure*}

Because the noise is independent between Fourier coefficients, the
log-likelihood of a rotation given one particle is a weighted least-squares
residual,
\begin{equation}
\ell(\Rot)\;=\;-\!\!\sum_{\mathbf{q}\in\Pi_{\Rot}}\!\!
\frac{\bigl\lvert y(\mathbf{q})-\CTF(\mathbf{q})\hat{\Vol}(\Rot^{-1}\mathbf{q})\bigr\rvert^{2}}
     {\sigma^{2}(k)} \;+\; \mathrm{const},
\label{eq:ell}
\end{equation}
and every estimator compared in this work is, explicitly or implicitly, an
attempt to maximize or to marginalize Eq.~\eqref{eq:ell}.

\subsection{Information geometry of orientation
estimation}\label{sec:methods:info}

Two questions decide whether one estimator can serve many specimens: how much orientation information a single particle carries, and how much of that information depends on which molecule is in the ice. Both follow from Eq.~\eqref{eq:ell}.

\paragraph{The derivative of a slice under rotation.}
Perturbations of $\Rot$ live in the Lie algebra $\mathfrak{so}(3)$. Take the
one-parameter family $\Rot(\delta)=\Rot\exp(\delta\,\hat{e}^{\wedge})$ for a
unit axis $\hat{e}\in\mathbb{S}^{2}$, an angle $\delta\in\mathbb{R}$ and the
skew-symmetric generator $\hat{e}^{\wedge}\in\mathfrak{so}(3)$. Writing
$\mathbf{p}=\Rot^{-1}\mathbf{q}$ and differentiating the sampled coefficient at
$\delta=0$,
\begin{equation}
\frac{\partial}{\partial\delta}\,
\hat{\Vol}\bigl(\Rot(\delta)^{-1}\mathbf{q}\bigr)\Big|_{\delta=0}
=-\,\nabla\hat{\Vol}(\mathbf{p})\cdot\bigl(\hat{e}\times\mathbf{p}\bigr),
\label{eq:dslice}
\end{equation}
whose second factor is the velocity of the sampling point under the rotation,
of magnitude
\begin{equation}
\bigl\lVert \hat{e}\times\mathbf{p}\bigr\rVert = k\,\sin\psi,
\qquad \cos\psi=\hat{e}\cdot\hat{\mathbf{p}} .
\label{eq:arc}
\end{equation}
A rotation by $\delta$ transports a component at radius $k$ along an arc of
length $k\delta\sin\psi$, so the same angular perturbation moves high-frequency
content proportionally further. This displacement is the geometric origin of
everything that follows, and the classical relation between angular accuracy
and attainable resolution rests on it \cite{rosenthal2003optimal}.

\paragraph{Orientation information grows as $k^{2}\,\mathrm{SSNR}(k)$.}

For complex Gaussian noise the Fisher information of $\delta$ is the
inverse-variance-weighted squared derivative of the model,
\begin{equation}
\mathcal{I}(\delta)
=2\!\!\sum_{\mathbf{q}\in\Pi_{\Rot}}\!\!
\frac{\CTF^{2}(\mathbf{q})}{\sigma^{2}(k)}\,
\bigl\lvert\nabla\hat{\Vol}(\mathbf{p})\cdot(\hat{e}\times\mathbf{p})\bigr\rvert^{2}.
\label{eq:fisher}
\end{equation}
Substituting Eq.~\eqref{eq:arc} and grouping the sum into shells gives the
contribution of shell $k$,

\begin{equation}
\mathcal{I}_k \;\propto\; k^{2}\,R_g^{2}\,
\underbrace{\frac{\CTF^{2}(k)\,P_k}{\sigma^{2}(k)}}_{\textstyle \mathrm{SSNR}(k)}\;n_k ,
\label{eq:k2law}
\end{equation}
in which $P_k$ is the signal power in shell $k$, $n_k\in\mathbb{N}$ the number of
Fourier samples it holds, $\mathrm{SSNR}(k)$ the spectral signal-to-noise ratio
(SSNR) of that shell, and $R_g$ the radius of gyration of the density about the
axis of rotation. Three reductions carry Eq.~\eqref{eq:fisher} to
Eq.~\eqref{eq:k2law}, each an approximation rather than an identity.
$\nabla\hat{\Vol}$ is the transform of $-2\pi i\,\mathbf{x}\Vol(\mathbf{x})$, so
its shell power is of order $(2\pi)^{2}R_g^{2}P_k$ rather than $P_k$; this
moment approximation is what carries the specimen's spatial extent into the
result. The gradient is then taken isotropic within a shell, which replaces the
projection onto one direction by
$\langle\lvert\nabla\hat{\Vol}\rvert^{2}\rangle\lVert\hat{e}\times\mathbf{p}
\rVert^{2}/3$ and leaves a constant from averaging $\sin^{2}\psi$. And $\Vol$ is
real, so summing over the full slice counts each independent coefficient twice,
a factor of about two absorbed into $n_k$.
The factor $R_g^{2}$ is not a nuisance constant: a larger particle carries its
high-frequency content through a longer arc for the same rotation, so
orientation is easier to determine. Read in real space, a point at radius $r$
moves by $r\delta$, and requiring that to stay within a resolution element $d$
gives $\delta\lesssim d/r$ --- the same inverse relation between angular
accuracy and attainable resolution that Rosenthal and Henderson write as
$\delta\lesssim d/D$ over a particle of diameter $D$ \cite{rosenthal2003optimal}.

Two multiplicities compound in Eq.~\eqref{eq:k2law}: the arc length grows as
$k$, contributing $k^{2}$ after squaring, and the shell population grows as
$n_k\propto k^{2}$. Orientation evidence is therefore concentrated at high
spatial frequency, where the SSNR is smallest. The useful band is the maximizer
of $k^{2}\,\mathrm{SSNR}(k)$, and that competition --- geometry pushing up,
radiation damage and the envelope pushing down --- sets a finite optimal
resolution range for pose assignment.

\paragraph{The prediction, and what the data give.}

Scoring each shell by a separately normalized correlation removes $n_k$ and the
shell's absolute power, leaving a curvature $\Lambda_k$ whose remaining specimen
dependence runs through the correlation $c_k$ that shell attains at the true
pose. It does not remove $R_g^{2}$, a property of the molecule rather than of
the shell; that factor is constant here because the exponent is fitted within a
single specimen, where it enters the intercept and not the slope. What remains
is a prediction with no free parameter,
\begin{equation}
\frac{\Lambda_k}{c_k}\;\propto\;k^{2}
\qquad\Longrightarrow\qquad
\frac{\mathrm{d}\log(\Lambda_k/c_k)}{\mathrm{d}\log k}\;=\;2 .
\label{eq:exponent}
\end{equation}
Measured on real particles at their deposited orientations --- the Fisher
information is defined at the true parameter --- the curvature follows
$\Lambda_k/c_k\propto k^{\,1.89}$ over shells $k=4$ to $32$, spanning $0.90$
decades (Fig.~\ref{fig:theory}A). Neighboring shells are correlated, and the fit
residuals carry a lag-one autocorrelation of $0.40$, so the ordinary
least-squares standard error of $0.037$ is optimistic; the
autocorrelation-robust value is $0.042$, placing the measurement $2.6\sigma$
below $2$. That deficit is not a bend in the power law, since adding a quadratic
term in $\log k$ improves the fit by $0.5\sigma$. Two effects depress the slope
uniformly and survive the $c_k$ normalization: a residual envelope or
$B$-factor, and noise bias in $c_k$ at the outer shells. We read the measurement
as agreement to within a few percent rather than exact confirmation. Either way
the scaling is obeyed by the data with no network involved.

\paragraph{Bound on attainable accuracy.}
Summing Eq.~\eqref{eq:k2law} over shells and inverting gives the
Cram\'er--Rao bound \cite{rao1945information,cramer1946mathematical} for an
unbiased estimator of a single rotation component,
\begin{equation}
\mathrm{Var}(\hat{\delta})\;\ge\;\mathcal{I}^{-1},
\qquad
\mathcal{I}\;=\;\kappa\,R_g^{2}\sum_{k}k^{2}\,\mathrm{SSNR}(k)\,n_k ,
\label{eq:crb}
\end{equation}
with $\kappa$ a purely geometric factor collecting the shell averages and
the Hermitian double-count above. Three consequences follow. Accuracy
improves through shells that carry signal, so band-limiting a particle
below the peak of $k^{2}\mathrm{SSNR}$ discards orientation information
irreversibly. The bound is per particle and independent of the estimator, so a
gap between Eq.~\eqref{eq:crb} and a measured error is a property of the
algorithm. And because $\mathrm{SSNR}$ is fixed by the sampling of a given
dataset, Eq.~\eqref{eq:crb} is the formal statement of the specimen-level
ceiling observed in Sec. (~\ref{sec:res:theory}).

Two qualifications set its scope. Equation~\eqref{eq:crb} bounds one rotation
component with the other two known, and is optimistic by a factor of order unity
against the corresponding element of the inverted $3\times3$ information matrix;
$\kappa$ also varies with the axis, since a rotation about the beam keeps the
sampling point in the slice plane at $\sin^{2}\psi=1$ while a perpendicular axis
averages to $\tfrac{1}{2}$, a factor $\sqrt{2}$ in angular standard deviation
that the three body-axis probes of Eq.~\eqref{eq:newton} see in aggregate. The
bound also governs the local basin around the true pose, and says nothing about
assigning a particle to the wrong basin. That second failure is a detection
question, set by whether the matched-filter statistic $\rho^{2}$ clears the
$2\ln\lvert\mathcal{G}\rvert$ threshold imposed by the competing candidates ---
the multiple-hypothesis accounting that 2D template matching applies to a
whitened matched filter searching orientations exhaustively against a reference
\cite{rickgauer2017single,lucas2021locating}. Neighboring grid poses give
correlated scores, so the effective number of independent candidates is smaller
than $\lvert\mathcal{G}\rvert$ and that threshold is conservative here. A local
precision bound and a global detection criterion together determine attainable
accuracy.

\paragraph{Why one estimator serves many specimens.}
Write the inference we wish to perform as a map from data and reference to a
distribution on rotations,
\begin{equation}
\mathcal{T}:\;(y,\Vol)\;\longmapsto\;p(\Rot\mid y,\Vol).
\label{eq:operator}
\end{equation}
Bayes' rule with Eq.~\eqref{eq:ell} gives
$p(\Rot\mid y,\Vol)\propto\exp\ell(\Rot;y,\Vol)$, so $\mathcal{T}$ is
determined by three objects: the transfer function, the noise spectrum, and the
slice geometry of the central-slice theorem. None of them depends on which
molecule is in the ice. The specimen enters Eq.~\eqref{eq:k2law} through
$P_k$ and its spatial extent $R_g$, both of which the reference supplies at
inference time.

A per-dataset estimator learns instead the partially applied map
$\mathcal{T}_{\Vol}: y\mapsto p(\Rot\mid y,\Vol)$ with $\Vol$ absorbed into the
parameters. The two differ in what generalizes. Fitting $\mathcal{T}_{\Vol}$
for $M$ specimens requires $M$ independent fits, each using only that
specimen's particles; fitting $\mathcal{T}$ once uses all particles from all
specimens for the single operator that is common to them, and treats $\Vol$ as
a nuisance argument redrawn at every training step: each step supplies one
reference volume together with a batch of $B$ particles generated from it, so a
single template bank is rendered once and shared by the whole batch
(Sec.~\ref{sec:methods:data}). Training across many
maps is then a variance-reduction device for the shared operator rather than a
compromise between specimens. 

\paragraph{Discretization floor.}
The posterior is evaluated on a finite grid $\mathcal{G}\subset\SO$, so even an
exact scorer inherits a quantization error. For $\lvert\mathcal{G}\rvert$ cells
covering $\SO$, whose Haar volume is $8\pi^{2}$ in the metric where geodesic
distance is the rotation angle, the mean nearest-neighbor spacing scales as
\begin{equation}
\Delta \;\approx\; \bigl(8\pi^{2}/\lvert\mathcal{G}\rvert\bigr)^{1/3},
\label{eq:spacing}
\end{equation}
giving $\SI{7.4}{\degree}$ at the $\lvert\mathcal{G}\rvert=36{,}864$ used here,
against a measured median nearest-neighbor spacing of $\SI{7.40}{\degree}$.
Reaching below this floor requires a continuous step, which motivates the
refinement of Sec.~(\ref{sec:methods:refine}); conversely, refining a grid whose
spacing is already below the Cram\'er--Rao scale of Eq.~\eqref{eq:crb} buys
nothing.

\begin{figure*}[t]
\centering
\includegraphics[width=\textwidth,height=0.24\textheight,keepaspectratio]{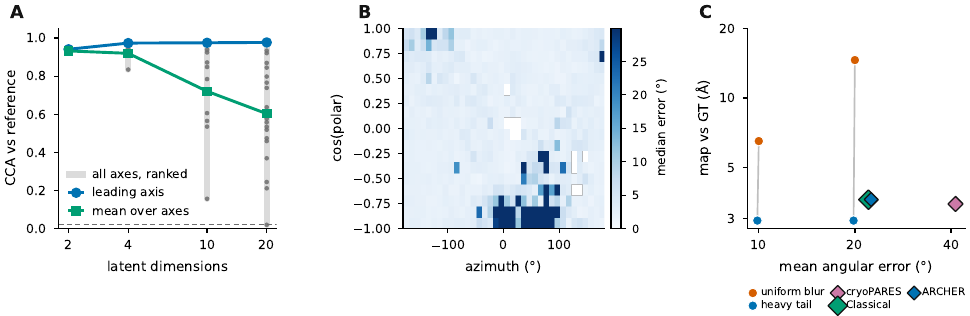}
\caption{\textbf{Error quantification from reconstruction maps.}
(A)~Canonical correlation against latent dimension on EMPIAR-10076: the leading
direction is flat while the mean declines, so it is the trailing directions that
decorrelate. (B)~Median error per cell of the view sphere on EMPIAR-10076.
(C)~Map resolution against \emph{mean} angular error, log--log. Circles are
manufactured perturbations at matched mean and opposite error shape --- at a
mean of $\SI{20}{\degree}$ the same particles give $\SI{2.94}{\angstrom}$ under
a heavy tail and $\SI{14.6}{\angstrom}$ under a uniform blur --- and diamonds are
the real estimators, which sit far to the right of those controls and far below
them. 
}
\label{fig:bench2}
\end{figure*}

\subsection{Reference-conditioned posterior}\label{sec:methods:model}

Templates $\mathbf{t}_i=\mathcal{F}^{-1}\Slice_{\Rot_i}\hat{\Vol}
\in\mathbb{R}^{L\times L}$ are rendered from the reference by central-slice
extraction on a HEALPix grid $\mathcal{G}\subset\SO$
\cite{gorski2005healpix,yershova2010generating} and encoded once per training
step; particles are encoded per image. Both encoders are residual convolutional
networks \cite{he2016deep} mapping into the unit sphere
$\mathbb{S}^{511}\subset\mathbb{R}^{512}$ ($10.4$~M parameters per branch): the
particle encoder $f_\theta:\mathbb{R}^{C\times L\times L}\to\mathbb{S}^{511}$
takes the $C$-channel image stack and the template encoder
$f_\phi:\mathbb{R}^{L\times L}\to\mathbb{S}^{511}$ a single rendered slice.
Pose logits are
\begin{equation}
s(\Rot_i \mid y) \;=\; \tau^{-1}\,\langle\, f_\theta(y),\; f_\phi(\mathbf{t}_i)\,\rangle ,
\qquad \Rot_i \in \mathcal{G},
\label{eq:logits}
\end{equation}
with $\tau>0$ a learned temperature
\cite{oord2018representation,radford2021learning} that falls from $0.07$ at
initialization to $0.0212$, sharpening the posterior by a factor of $3.3$. The
objective is cross-entropy against a geodesic soft target
$q_i \propto \exp(-\geo(\Rot_i,\Rot^\star)^2/2\sigma_q^2)$ with
$\sigma_q=\SI{6}{\degree}$, which supplies gradient to neighboring cells and
respects the metric of $\SO$. Encoder topology, the hard-negative pool and the
optimizer settings are given in App.~\ref{app:arch}.

The CTF enters as image content: the $C$
channels carry the raw particle and its phase-flipped counterpart, and templates
are rendered without a transfer function. This costs information and buys
amortization, and both sides are quantifiable. The sufficient statistic of
Eq.~\eqref{eq:ell} is $\langle\CTF\cdot\mathbf{t},\,y\rangle/\sigma^{2}$, which
weights each shell by the transfer function's amplitude; phase flipping keeps
the sign and discards that weighting. Writing $\langle\cdot\rangle_w$ for
the average under the weight $w_k=P_k/\sigma^{2}(k)$, the ratio of the two
informations is

\begin{equation}
\zeta=\frac{\mathcal{I}_{\mathrm{flip}}}{\mathcal{I}_{\mathrm{matched}}}
=\frac{\langle\lvert\CTF\rvert\rangle_w^{2}}{\langle\CTF^{2}\rangle_w}
=\biggl[1+\frac{\mathrm{Var}_w(\lvert\CTF\rvert)}
                {\langle\lvert\CTF\rvert\rangle_w^{2}}\biggr]^{-1},
\label{eq:flipeff}
\end{equation}
so the loss is the relative variance of $\lvert\CTF\rvert$ across the band, and
$\zeta\le1$ by Cauchy--Schwarz with equality for a flat transfer function.

Evaluated over the real per-particle defocus values of EMPIAR-10409 at the
measured noise spectrum, $\zeta=0.542\pm0.009$: the phase-flipped scorer retains
$54\%$ of the available orientation information, a factor $1.36$ in angular
standard deviation. What it buys is that templates are encoded once per specimen
rather than once per particle, since applying $\CTF$ inside the similarity would
make the template embedding defocus-dependent. That is the difference between
one template bank per specimen and one per particle, and it is what makes
amortization affordable.

\subsection{Diffusion denoising channel}\label{sec:methods:diffusion}

At the signal-to-noise ratios of a single particle, the shells that carry
orientation information by Eq.~\eqref{eq:k2law} are the ones the noise
dominates, so restoring amplitude there is the operation that stands to sharpen
a pose. The diffusion channel performs it: a Fourier-domain denoiser, trained
jointly with the pose objective, supplies a denoised copy $\mathrm{DN}(y)$ of
the particle as a third input channel to the encoder. The construction turns on
one observation. In a diffusion model the data sit at $t=0$ and noise is added
to reach the terminal state; here we choose a forward process whose terminal
state \emph{is} the measurement, by interpolating the identity into the transfer
function,

\begin{equation}
\begin{split}
y_t &\;=\; \mathbf{A}_t\,\hat{y}_0 \;+\; \varepsilon_t ,\\
\mathbf{A}_t &= (1-w_t)\,\mathbf{1} + w_t\,\CTF ,
\qquad w_t = \tfrac{t}{T-1},
\end{split}
\label{eq:forwardproc}
\end{equation}
where $t\in\{0,\dots,T-1\}$ indexes the noise level, $T\in\mathbb{N}$ is the
number of levels, $\hat{y}_0\in\mathbb{C}^{L\times L}$ is the clean
Fourier-domain image, $\mathbf{A}_t\in\mathbb{R}^{L\times L}$ is the transfer
operator acting elementwise, $\mathbf{1}\in\mathbb{R}^{L\times L}$ is the
all-ones array, $w_t\in[0,1]$ is the interpolation weight, and
$\varepsilon_t\in\mathbb{C}^{L\times L}$ is zero-mean Gaussian of variance
$(1-\bar\alpha_t)\,\sigma^{2}(k)$, with $\bar\alpha_t\in[0,1]$ following a cosine
schedule \cite{ho2020ddpm,nichol2021improved}. Because a recorded particle is
formed as $\CTF\cdot\hat{y}_0+\eta$, Eq.~\eqref{eq:forwardproc} places it at
$t=T-1$ \emph{exactly}: the denoiser is evaluated once at a known level rather
than sampled over a reverse trajectory, which is what makes it cheap enough to
sit inside the training loop and differentiable through it. Equation~\eqref{eq:forwardproc} carries no
$\sqrt{\bar\alpha_t}$ factor on the signal and its terminal marginal is the
measurement rather than a fixed Gaussian, which places it in the
corruption-schedule family, where the forward process is a physical degradation
and the terminal state is the observation
\cite{bansal2023cold,daras2023soft,delbracio2023inversion,liu2023i2sb,luo2023image}.

A U-Net $\varepsilon_\vartheta$ \cite{ronneberger2015unet}, with parameters
$\vartheta$ distinct from those of the two encoders, predicts the noise from the
real and imaginary parts of $y_t$, the transfer function and a sinusoidal time
embedding. It is trained with the $\varepsilon$-prediction objective
\cite{ho2020ddpm}, weighted by the inverse noise variance so that every shell
contributes on the scale of its own uncertainty,
\begin{equation}
\mathcal{L}_{\varepsilon}
=\E_{t,\varepsilon}\!\left[\;\sum_{k}
\frac{\bigl\lVert \varepsilon(k)-\varepsilon_\vartheta\bigl(y_t,t,\CTF\bigr)(k)\bigr\rVert^{2}}
{(1-\bar{\alpha}_t)\,\sigma^{2}(k)}\right],
\label{eq:epsloss}
\end{equation}
and the total objective is $\mathcal{L}=\mathcal{L}_{\mathrm{CE}}
+\mathcal{L}_{\varepsilon}$, so two gradients reach this network: the
$\varepsilon$-prediction loss with $t$ drawn uniformly, and the pose
cross-entropy through the pipeline. Given $\hat{\varepsilon}$, the clean estimate
follows by Wiener inversion \cite{wiener1949extrapolation},
\begin{equation}
\hat{y}_0 \;=\; \bigl(y_t-\hat{\varepsilon}\bigr)\,
\frac{\mathbf{A}_t}{\mathbf{A}_t^{2}+\lambda},
\label{eq:wiener}
\end{equation}
with $\lambda=\sigma_n^{2}/\sigma_s^{2}>0$ the ratio of leftover-noise to
signal power. The gain is the linear estimator
$\hat{y}_0=\gamma\,(y_t-\hat{\varepsilon})$ minimizing
$\E\lvert\hat{y}_0-y_0\rvert^{2}$ when the residual uncertainty is treated as a
signal of prior power $\sigma_s^{2}$ against leftover noise of power
$\sigma_n^{2}$, which gives
$\gamma^{\star}=\mathbf{A}_t/(\mathbf{A}_t^{2}+\lambda)$. The regularizer is
what makes Eq.~\eqref{eq:wiener} usable: direct inversion by $\mathbf{A}_t$
diverges at the zeros of $\CTF$, whereas this gain is bounded by
$1/(2\sqrt{\lambda})$ at $\mathbf{A}_t=\sqrt{\lambda}$ and falls to zero as
$\mathbf{A}_t\to0$ \cite{wiener1949extrapolation}, so the estimator declines to
invent content in the transfer function's blind bands.

The channel acts on fine
precision. Over $150$
step-matched held-out evaluations on synthetic proteins, the median angular
error is $5.07^\circ$ against $5.22^\circ$ and the fraction within
$\SI{5}{\degree}$ is $0.493$ against $0.474$, leading at $122$ and $121$ of the
$150$ evaluations, while the fraction within $\SI{15}{\degree}$ is unchanged at
$0.755$ against $0.753$: the denoised channel sharpens the estimate inside the
correct basin and preserves the coarse assignment. On experimental particles, we reached pose median on EMPIAR-10409 to
$\SI{0.01}{\degree}$ and the reconstructions to $\SI{0.001}{\angstrom}$. The
results reported throughout this work are from the three-channel configuration,
CTF-modulated particles, phase-flipped, and the denoising channel.

\subsection{Hierarchical refinement}\label{sec:methods:refine}

Inference proceeds from the grid maximum to a continuous estimate. Given
current poses, a volume is rebuilt by Wiener-filtered backprojection, every
particle is re-scored against re-rendered templates by a shell-normalized
correlation restricted to low shells, and each pose is polished by a
derivative-free Newton step.

Let $s(\Rot)$ be the score of Eq.~\eqref{eq:logits} or its classical
counterpart. Parameterizing a neighborhood of the current estimate by the
exponential map, $\Rot(\boldsymbol{\delta})=\Rot\exp\bigl(\sum_j\delta_j
\hat{e}_j^{\wedge}\bigr)$ with $\boldsymbol{\delta}\in\mathbb{R}^{3}$, a
second-order expansion gives
\begin{equation}
s(\boldsymbol{\delta})\;\simeq\;s(\mathbf{0})
+\mathbf{g}^{\!\top}\boldsymbol{\delta}
+\tfrac{1}{2}\,\boldsymbol{\delta}^{\!\top}\mathbf{\Lambda}\boldsymbol{\delta},
\qquad
\mathbf{g}=\nabla_{\boldsymbol{\delta}}s,\;\;
\mathbf{\Lambda}=\nabla^{2}_{\boldsymbol{\delta}}s ,
\label{eq:expansion}
\end{equation}
with gradient $\mathbf{g}\in\mathbb{R}^{3}$ and Hessian
$\mathbf{\Lambda}\in\mathbb{R}^{3\times3}$, whose stationary point is
$\boldsymbol{\delta}^{\star}=-\mathbf{\Lambda}^{-1}\mathbf{g}$. The update is
applied on the group, $\Rot\leftarrow\Rot\exp(\boldsymbol{\delta}^{\star\wedge})$,
so the estimate never leaves $\SO$.

We evaluate $\mathbf{g}$ and the diagonal of $\mathbf{\Lambda}$ by central
differences, which needs no gradient through the scorer, so the same routine
refines the learned posterior and the classical matched filter. Writing
$s^{\pm}_j=s\bigl(\Rot\exp(\pm\epsilon\,
\hat{e}_j^{\wedge})\bigr)$ and $s^{0}=s(\Rot)$ for the score probed along the
three body axes $\hat{e}_j\in\mathbb{S}^{2}$, $j=1,2,3$, at probe angle
$\epsilon>0$,
\begin{equation}
g_j=\frac{s^{+}_j-s^{-}_j}{2\epsilon},
\qquad
\Lambda_j=\frac{s^{+}_j+s^{-}_j-2s^{0}}{\epsilon^{2}},
\label{eq:newton}
\end{equation}
are the directional first and second derivatives, and the update
$\Rot \leftarrow \Rot\exp(\boldsymbol{\delta}^{\wedge})$ with
$\delta_j=g_j/\lvert \Lambda_j\rvert$ is taken where $\Lambda_j<0$ and
clamped to $\SI{1}{\degree}$ per axis otherwise. Where the curvature has
the wrong sign the step reverts to that clamped gradient move, which keeps the
update stable on the flat plateaus between grid cells. The probe scale is set by
the two angular scales the problem supplies: it must exceed the noise-induced
roughness of the score and stay below the curvature scale of its peak, and
$\epsilon=\SI{1}{\degree}$ against a grid spacing of $\SI{7.4}{\degree}$
satisfies both (Fig.~\ref{fig:theory}C). The diagonal of the Hessian is
evaluated at six renders per step; the full $3\times3$ would cost six more.

Two loops are nested here and are counted separately in Alg.~\ref{alg:infer}.
The \emph{outer} loop rebuilds the volume, re-scores against it and refines, and
runs $T_{\mathrm{iter}}=3$ times; the \emph{inner} Newton loop takes
$T_{\mathrm{newton}}=4$ update steps within each of those, at fixed volume and
fixed templates.
Half-sets are processed independently throughout --- each half is backprojected
into its own volume and every particle is scored only against the volume built
from the opposite half --- preserving the gold-standard separation
\cite{scheres2012goldstandard,henderson2012outcome}.

\begin{figure}[t]
\centering
\includegraphics[width=\linewidth,height=0.35\textheight,keepaspectratio]{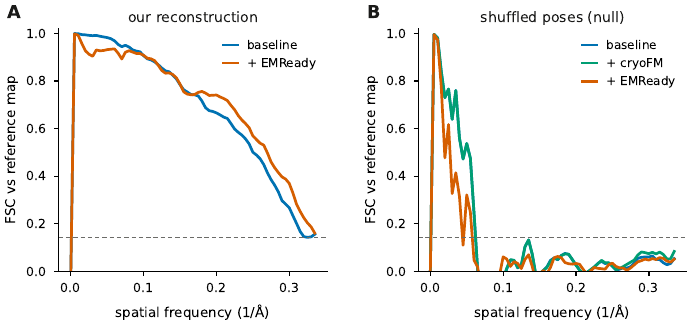}
\caption{\textbf{Map restoration.} Shell correlation before and after
restoration.}
\label{fig:restore}
\end{figure}

\begin{algorithm}[t]
\begin{algorithmic}[1]
\State \textbf{input} Reference Volume Bank $\{\Vol^{(1)},\dots,\Vol^{(M)}\}$, Grid $\mathcal{G}$ \\ 
\For{each training step}
  \State sample a specimen $\Vol\sim$ bank and a noise level $\mathrm{SNR}$
  \State sample true rotations $\{\Rot^\star_b\}_{b=1}^{B}$ uniformly on $\SO$
  \State sample defocus $\{\xi_b\}_{b=1}^{B}$ from an experimental star file
         \Comment{one per particle}
  \State $\CTF_{\xi_b}\gets$ transfer function of $\xi_b$ on this specimen's grid
  \State $y_b \gets \mathcal{F}^{-1}[\CTF_{\xi_b}\cdot\Slice_{\Rot^\star_b}\hat{\Vol}] + \eta_b$
         \Comment{Eq.~\eqref{eq:forward}}
  \State $\mathcal{K}\gets$ uniform cells $\cup$ mined hard negatives $\cup\;\{\Rot^\star_b\}$
  \State $\mathbf{t}_i \gets \mathcal{F}^{-1}\Slice_{\Rot_i}\hat{\Vol}$ for $\Rot_i\in\mathcal{K}$;\;\;
         $v_i \gets f_\phi(\mathbf{t}_i)$ \Comment{encoded once per step}
  \State $u_b \gets f_\theta\bigl(\mathrm{stack}(y_b,\;\mathrm{PF}(y_b))\bigr)$
         \Comment{third channel $\mathrm{DN}(y_b)$}
  \State $s_{bi} \gets \tau^{-1}\langle u_b, v_i\rangle$ \Comment{Eq.~\eqref{eq:logits}}
  \State $q_{bi} \propto \exp\bigl(-\geo(\Rot_i,\Rot^\star_b)^2/2\sigma_q^2\bigr)$
  \State $\mathcal{L}\gets \mathrm{CE}(s_b,q_b)\;\,+\,\mathcal{L}_{\varepsilon}\,$
         \Comment{ Eq.~\eqref{eq:epsloss}}
  \State update $\theta,\phi,\tau$ and the denoiser
\EndFor
\end{algorithmic}
\algcaption{Training. One step is one specimen; the reference is data rather
than a parameter, so a single set of weights is fit across the bank.}
\label{alg:train}
\end{algorithm}

\begin{algorithm}[t]
\begin{algorithmic}[1]
\State \textbf{input} Particles $\{y_b\}$ split into halves $h\in\{0,1\}$,
       Predicted target volume $\hat{\Vol}^{(0)}$, Grid $\mathcal{G}$ \\ 
\State $v_i\gets f_\phi(\mathcal{F}^{-1}\Slice_{\Rot_i}\hat{\Vol}^{(0)})$ for all $\Rot_i\in\mathcal{G}$
\State $\Rot_b^{(0)}\gets \argmax_{\Rot_i\in\mathcal{G}}\;\tau^{-1}\langle f_\theta(y_b),v_i\rangle$\\ 
\For{$m=0,\dots,T_{\mathrm{iter}}-1$}\Comment{$T_{\mathrm{iter}}=3$}
  \For{$h=0,1$}\Comment{separate half-sets}
    \State $\hat{\Vol}^{(m+1)}_h\gets
           \mathrm{backproject}\bigl(\{y_b,\Rot^{(m)}_b,\CTF_b\}_{b\in h}\bigr)$
           \Comment{Wiener filtered}
  \EndFor
  \State $\Rot^{\prime}_b\gets \argmax_{\Rot_i\in\mathcal{G}}\;
         \mathrm{corr}_{\mathrm{shell}}\bigl(y_b,\Slice_{\Rot_i}\hat{\Vol}^{(m+1)}_{1-h(b)}\bigr)$
         \Comment{opposite half}
  \State $\Rot^{(m+1)}_b\gets \textproc{NewtonRefine}\bigl(\Rot^{\prime}_b\bigr)$
         \Comment{Eq.~\eqref{eq:newton}}
\EndFor
\Statex
\Function{NewtonRefine}{$\Rot$}
  \For{$n=1,\dots,4$}
    \For{$j=1,2,3$}
      \State $s^{\pm}_j\gets s\bigl(\Rot\exp(\pm\epsilon\,\hat{e}_j^{\wedge})\bigr)$
      \State $g_j\gets\frac{s^{+}_j-s^{-}_j}{2\epsilon}$;\quad
             $\Lambda_j\gets\frac{s^{+}_j+s^{-}_j-2s^{0}}{\epsilon^{2}}$
      \State $\delta_j\gets g_j/\lvert \Lambda_j\rvert$ if $\Lambda_j<0$ else clamped $g_j$
    \EndFor
    \State $\Rot\gets\Rot\exp(\boldsymbol{\delta}^{\wedge})$
  \EndFor
  \State \Return $\Rot$
\EndFunction
\end{algorithmic}
\algcaption{Hierarchical inference. Half-sets are processed independently, so
no particle contributes to the reference against which it is scored.}
\label{alg:infer}
\end{algorithm}

\begin{figure*}[!ht]
\centering
\includegraphics[width=\textwidth,height=0.55\textheight,keepaspectratio]{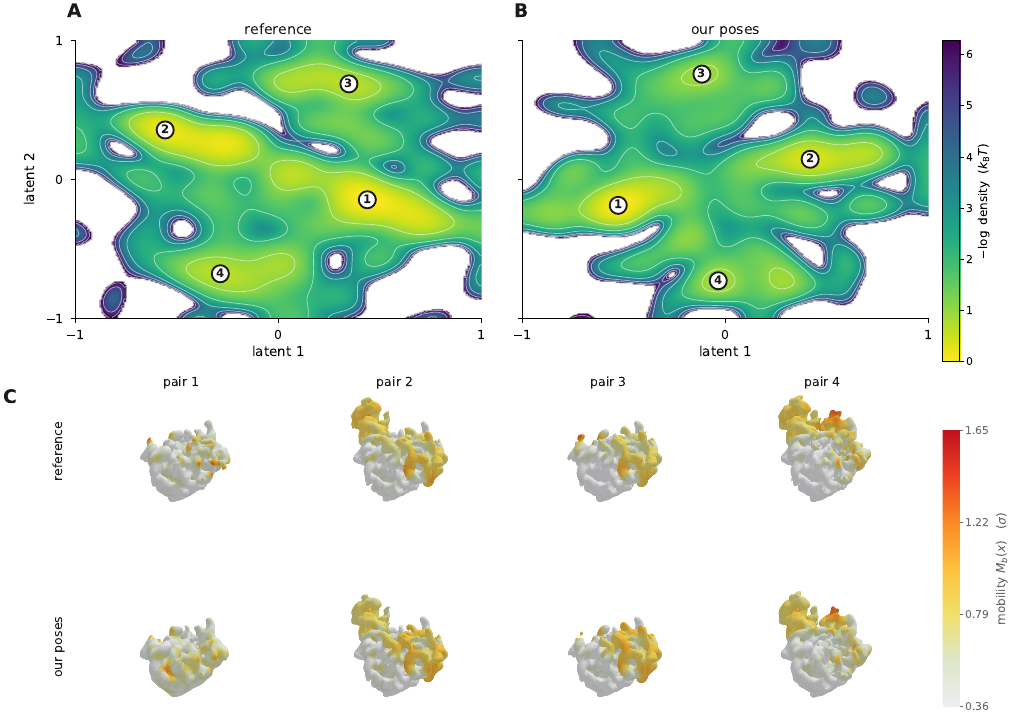}
\caption{\textbf{Conformational landscape.} (A,~B)~Deconvolved conformational
density over the leading two latent dimensions for the same 91{,}899
EMPIAR-10076 particles embedded twice, with the deposited poses and with ours,
plotted as $-\log p$ in units of $k_\mathrm{B}T$. Both panels use the same deconvolution weight, and
each latent axis is normalized by its own bounds; markers are basins.
(C)~The reconstruction at each basin, colored by the mobility of
Eq.~\eqref{eq:mobility} in units of the map's own standard deviation. Columns
pair basins by map correlation: the two latent bases are independent, so basin
$i$ of one panel need not be basin $i$ of the other.}
\label{fig:land}
\end{figure*}

\subsection{Data}\label{sec:methods:data}

\emph{Volumes and Particles.} Training particles are sampled from the training datasets (pertaining to different structures) using the known pose information relative to each structure. Our training sets are from the Electron Microscopy Data Bank (EMDB) \cite{lawson2016emdatabank}, where we used $3{,}330$ structures while $100$ structures are held out for evaluation. For each volume of each structure, \textit{training episodes} are drawn from those $3{,}330$ structures during training. Here, a training episode is a volume and $96$ particles generated through the forward model of Sec.~(\ref{sec:methods:forward}): a central slice of the volume, a CTF whose defocus is drawn from the corresponding training set, and additive noise colored by a power spectral density measured from real micrographs rather than white, at signal-to-noise ratios sampled from $0.005$ to $0.075$. Training runs for $2\times10^{6}$ of these episodes.

Moreover, for the volumes, we used the deposited half-maps, whose native boxes span $94^3$ to $268^3$ at roughly $\SI{1.5}{\angstrom}$ per voxel. Each half-map is reduced to a common $64^3$ box by cropping in Fourier space rather than in real space: a real-space crop would truncate the larger proteins, whereas discarding high frequencies retains the whole molecule at coarser sampling, which is the operation the resolution argument of Sec.~(\ref{sec:methods:info}) assumes. Because the crop is to a fixed box rather than a fixed sampling, voxel size varies with the structure (with median value of $\SI{4.7}{\angstrom}$ alongside $90\%$ of structures between $3.1$ and $\SI{6.9}{\angstrom}$), so the model meets a range of scales during training.



\subsection{Evaluation}\label{sec:methods:eval} 

The held out structures are scored on simulated particles, which isolates generalization to unseen proteins from the change of domain. Experimental performance is measured, without retraining, on deposited data from the Electron Microscopy Public Image Archive \cite{iudin2016empiar}: EMPIAR-10076 for pose accuracy and heterogeneity, and the CESPED benchmark targets \cite{sanchezgarcia2023cesped} EMPIAR-10166, 10280, 10409, 10648 and 11120 for reconstruction. Training volumes come from the EMDB and evaluation particles from EMPIAR, but the deposited structures of the benchmark targets are not among the volumes trained on. Meanwhile, the inference pipeline, detailed in Alg.~(\ref{alg:infer}), takes in the predicted volume of the target structure $\hat{\Vol}^{(0)}$, and iteratively refines the pose information (to refine the volume) for $T_\text{iter}$ iterations.


\begin{figure*}[!ht]
\centering
\includegraphics[width=\textwidth,height=0.42\textheight,keepaspectratio]{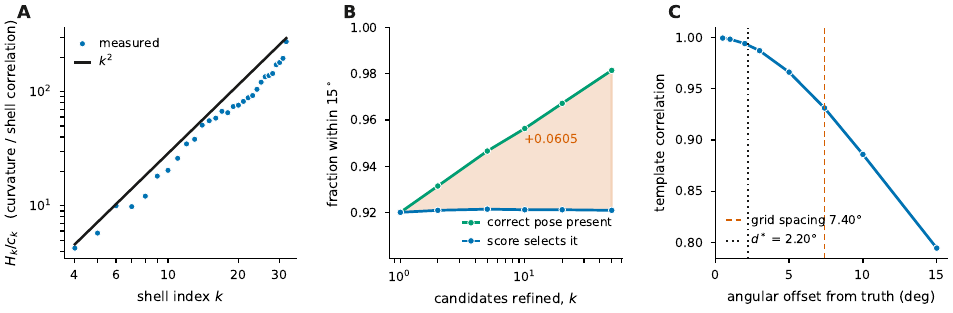}
\caption{\textbf{Information, selection and scale.} (A)~Per-shell curvature
normalized by the correlation $c_k$ that shell attains at the true pose,
against shell radius, with the $k^{2}$ prediction of Eq.~\eqref{eq:k2law}. The
measured exponent over $k=4$--$32$ is $1.89$, with an ordinary least-squares
standard error of $0.037$ and an autocorrelation-robust one of $0.042$; the
latter is the one to read, since neighboring shells are correlated.
(B)~Fraction of particles for which a correct pose is
present among refined candidates, against the fraction for which the score
selects it. (C)~The score landscape around the true pose, showing the grid
spacing and the curvature scale.}
\label{fig:theory}
\end{figure*}

Angular error is the geodesic distance on $\SO$ to the deposited pose, reported as a median and as the fraction within a threshold. Map metrics follow the CESPED protocol \cite{sanchezgarcia2023cesped}: half-sets are  reconstructed independently with RELION \cite{kimanius2021relion4} at the predicted orientations and compared to the deposited structure inside the benchmark mask, with resolution read at the $0.143$ criterion \cite{rosenthal2003optimal}. Conformational agreement is the canonical correlation between RECOVAR embeddings \cite{gilles2025recovar} of identical particles differing only in rotations, with a shuffled-pose floor measured through the identical pipeline. Restoration is evaluated with EMReady \cite{he2023emready} and cryoFM \cite{zhou2025cryofm} using their published metrics. See Figs.~(\ref{fig:bench}) and (\ref{fig:bench2}) for the benchmark results.

Conformational landscapes are compared through those same embeddings. RECOVAR's
deconvolved density $p(z)$ \cite{gilles2025recovar} reads as a free energy through
$-\log p(z)$, and the local maxima of $p$ are the \emph{basins}, the stable states.
Comparing the map at one basin with all the others would treat a difference across a
high barrier the same as one the particle crosses freely, so each destination is
weighted by how likely the particle is to reach it,
\begin{equation}
\begin{split}
M_b(x) &= \frac{\sum_{b'\neq b} w_{b\to b'}\bigl|V_{b'}(x)-V_b(x)\bigr|}
            {\sum_{b'\neq b} w_{b\to b'}}, \\
w_{b\to b'} &= e^{-[E^\ddagger(b,b')-E_b]/k_\mathrm{B}T},
\end{split}
\label{eq:mobility}
\end{equation}
with $V_b$ the map at basin $b$ and $E^\ddagger$ the lowest high point on any path
joining two basins --- the minimax barrier used to organize energy landscapes into
disconnectivity trees \cite{becker1997topology}, with the exponential weight the
Arrhenius factor for escape over it \cite{kramers1940brownian}. This \emph{mobility}
is large where a structure differs from the states the particle can actually reach.
Two pose sets are compared by matching their basins on map correlation and then
comparing the mobility fields at the matched basins.

\section{Results}\label{sec:results}

\subsection{Pose information prediction}\label{sec:res:amortization}

\methodname{} embeds a particle and a bank of reference templates into a
shared unit sphere and reads the pose off the maximum of a temperature-scaled cosine similarity (see Fig.~\ref{fig:arch} and Alg.~\ref{alg:train}). The rotation grid, for the reference volume, is a HEALPix discretization of $\SO$ \cite{gorski2005healpix,yershova2010generating} with
$36{,}864$ cells at a median spacing of $7.4^\circ$.

For the evaluation set of $100$ structures, we evaluated on $768$ simulated particles per structure at the signal-to-noise $0.02$--$0.30$ on the order-3 grid. \methodname{} reaches a median angular error of $\SI{5.0}{\degree}$ and places $77\%$ of particles within $\SI{15}{\degree}$. It selects the exactly correct grid cell, one of $36{,}864$, for $41\%$ of particles (shown in Fig.~\ref{fig:bench}).

Moreover, transferability of \methodname{} to experimental particles holds without retraining. When applied across all five CESPED targets (Fig.~\ref{fig:bench}A), EMPIAR-10280 and 10648 carry point-group symmetry, and  rotating such a particle
onto a symmetry-related orientation reproduces the same
projection image pixel for pixel. Relying on symmetry based on the reference volumes, we minimize the geodesic over the group. The D$_2$ target reads initially at $\SI{166.8}{\degree}$ and after utilizing symmetry-aware, at $\SI{9.1}{\degree}$. The correction to D$_2$ leaves the three C$_1$ targets bit-identical and does not rescue random poses, which stay at $\SI{90}{\degree}$. Detecting initial symmetry point-group

On EMPIAR-10076, a mixture of assembly intermediate  of the \emph{E.\ coli} large ribosomal subunit, \methodname{} attains a median error of $\SI{2.5}{\degree}$ with $78\%$ of particles within $\SI{5}{\degree}$ of the deposited pose (Fig.~\ref{fig:s1}). The learned classifier's grid argmax gives $\SI{6.16}{\degree}$. Re-scoring against the reference with a shell-normalized matched filter gives $\SI{4.53}{\degree}$, and the curvature step gives $\SI{2.54}{\degree}$. Since the refinement step searches the whole grid, the ground-truth precision is reached by the classical stage (Fig.~\ref{fig:bench}A).

We ran cryoPARES \cite{sanchezgarcia2025cryopares} on the four targets by first fitting to each of the targets and scored on the particles it did not train on (Fig.~\ref{fig:bench}A and Table~\ref{tab:cpunseen}).  Median error across them is $9.4$ for the network, $7.5$ for the full pipeline, $6.5$ classical and $\SI{4.2}{\degree}$ for cryoPARES (Fig.~\ref{fig:bench}A). It is the more accurate estimator wherever it can be applied, and that qualification is the whole of the difference between the two methods. The precision did not stay the same, once we reduced the fraction of particles placed within $\SI{15}{\degree}$ for fitting. Here, the classical method leads at $0.818$, followed by cryoPARES at $0.805$ and our pipeline at $0.718$. cryoPARES is more precise when it fits on abundance of particles per structure. Median and tail measure different things here, and we report both in Table~(\ref{tab:cpunseen}). 

Since cryoPARES is fit to one specimen at a time, we measured the performance on untrained targets. Applying each of the four trained models to each of the four targets, on the same particles and with the network stage alone, gives a median of $\SI{3.9}{\degree}$ to $\SI{8.8}{\degree}$ on the diagonal and $\SI{80}{\degree}$ to $\SI{135}{\degree}$ everywhere else (see Table~\ref{tab:transfer}). The off-diagonal values sit at the chance level of each target --- $\SI{120}{\degree}$ at C$_1$, $\SI{97}{\degree}$ at C$_2$, $\SI{90}{\degree}$ at D$_2$. The comparison in Fig.~\ref{fig:bench}A displays the model excellent on the single structure it is fit to, but not transferable to the others.

Reference-free estimators solve a harder problem -- no reference, no
translations -- and are compared separately in the supplementary Fig.~(\ref{fig:s3}) rather than alongside here. We modulated input data size for CryoFastAR on a homogeneous target in the supplementary Fig.~(\ref{fig:s2}). They are not weak baselines when given enough data. cryoDRGN's ab-initio schedule sits at chance on the $8{,}000$ particles these arms are scored on, but at $100{,}000$ reaches $1.96$, $5.32$ and $\SI{17.09}{\degree}$ on the three C$_1$ targets, ahead of our network on each; it degrades on the two symmetric targets, for which its homogeneous entry point offers no symmetry option. Their for these reference-free models incur more inference cost and risks poor transferability.

Errors concentrate in a band of viewing directions rather than spreading over
the sphere (Fig.~\ref{fig:bench2}B). The residual error of \methodname{} is localized
in orientation rather than diffusive, which is what makes it addressable by better
candidate selection. Furthermore, an uneven distribution of viewing directions attenuates resolution on its own, through a sampling compensation factor that weights each direction by how often it is occupied \cite{baldwin2020non}. 

\begin{figure}[!ht]
\centering
\includegraphics[width=\linewidth,height=0.30\textheight,keepaspectratio]{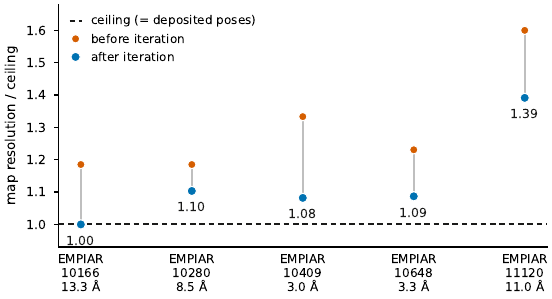}
\caption{\textbf{Cross-target reconstruction against the sampling bound.}
Five CESPED targets (EMPIAR-10166, 10280, 10409, 10648, 11120), each evaluated
against the Nyquist limit of of the $64^{3}$ box the pose estimate is made in. The y-axis is the ratio of map resolution to that ceiling, before and after hierarchical refinement.}
\label{fig:cross}
\end{figure}

\subsection{Reconstructions quality}\label{sec:res:maps}

Pose accuracy matters mainly through the map it produces. We evaluate with the CESPED benchmark protocol \cite{sanchezgarcia2023cesped}, which reconstructs each half-set independently at the predicted orientations and scores the result against the deposited structure.

On EMPIAR-10409, \methodname{} reconstructs to $\SI{3.634}{\angstrom}$ against the deposited map. The classical matched filter reaches $\SI{3.625}{\angstrom}$ and cryoPARES \cite{sanchezgarcia2025cryopares} $\SI{3.477}{\angstrom}$: all three lie within $\SI{0.16}{\angstrom}$ of one another, and their half-map resolutions span $\SI{0.05}{\angstrom}$ (Fig.~\ref{fig:bench}C). On this measure \methodname{} is comparable in reconstruction quality to the supervised method.

The scoring is not uniform across the benchmark. Scored the same way on all
four targets, our reconstructions trail cryoPARES by $\SI{0.16}{\angstrom}$ and $\SI{0.22}{\angstrom}$ on EMPIAR-10409 and 10648 but by $\SI{2.11}{\angstrom}$ and $\SI{2.06}{\angstrom}$ on EMPIAR-10166 and 10280 (Table~\ref{tab:maptargets}). Feeding more particles improved our reconstruction quality. We reached cryoPARES's quality when supplementing $203{,}000$ and $117{,}478$ on EMPIAR-10409 and EMPIAR-10648.

\begin{figure*}[!ht]
\centering
\includegraphics[width=\textwidth,keepaspectratio]{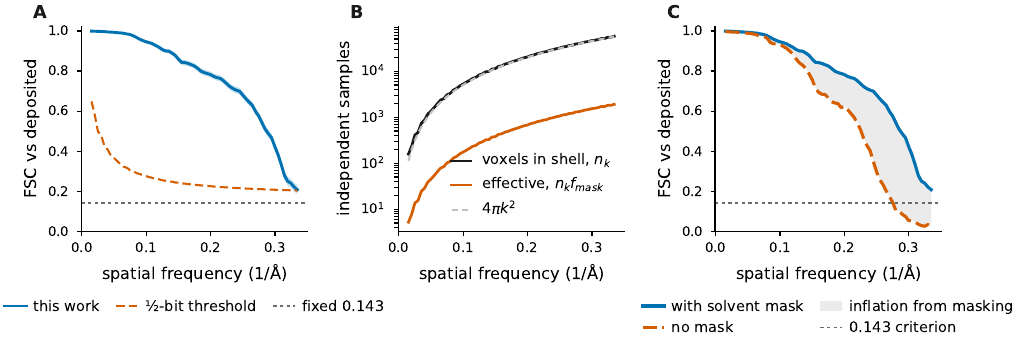}
\caption{\textbf{Shell correlation carries its own statistics.}
All three panels are the \methodname{} reconstruction of EMPIAR-10409 scored
against the deposited map, at the native $\SI{1.47}{\angstrom}$ sampling.
(A)~The reconstruction against the fixed $0.143$ criterion and against the
$\tfrac{1}{2}$-bit criterion, which scales with the number of samples in the
shell \cite{vanheel2005fourier}; the two disagree most where samples are
fewest. (B)~Samples per shell, following $4\pi k^2$, and the effective count
after masking, which is lower by the mask's volume fraction ($3.3\%$ here).
(C)~The same reconstruction scored masked and unmasked: masking raises the
curve throughout, and here pushes the $0.143$ crossing past Nyquist altogether,
so the shift in reported resolution is a lower bound of
$\SI{0.63}{\angstrom}$ rather than a measured value. The classical arm and
cryoPARES are censored at the same $\SI{2.94}{\angstrom}$, so on this target
the masked criterion cannot separate the three.}
\label{fig:physics}
\end{figure*}

Now, we questioned whether the resolution of reconstruction could reach any more meaningful levels. On EMPIAR-10409 the masked shell correlation never falls to $0.143$ within the band (Fig.~\ref{fig:physics}C). All reconstructions of classical, cryoPARES, and \methodname{}  are already at  the $\SI{2.94}{\angstrom}$ Nyquist of the native sampling for EMPIAR-10409 and $\SI{2.99}{\angstrom}$ for EMPIAR-10648 (Table~\ref{tab:maptargets}). Therefore, a small difference in the resolution score does not guarantee a better map. Fig.~(\ref{fig:physics}) shows how far that score moves under masking alone. 

On five CESPED targets spanning box sizes from $136$ to $284$ voxels and both $200$ and $300$~kV optics, \methodname{} reproduces the deposited density on every one of them (Fig.~\ref{fig:maps}A). Embedded in a common principal-component space, the five targets occupy distinguishable regions (Fig. ~\ref{fig:maps}B). Although data collection differs in defocus, contrast and sampling, the molecules in feature space are relatively close.  Measured instead against the Nyquist limit of the $64^{3}$ working box the estimate is made in, the reconstructions sit within a factor of $1.0$--$1.4$ of that limit on all five (Fig.~\ref{fig:cross}). The map quality reported above for EMPIAR-10409 ($\SI{3.63}{\angstrom}$) is measured at the native $\SI{1.47}{\angstrom}$ sampling.


\paragraph{Error shape governs the map.}
What limits a reconstruction is not the average pose error but its shape.
Perturbing the deposited poses by a uniform $\SI{10}{\degree}$ blur degrades the
map to $\SI{6.5}{\angstrom}$, while a heavy-tailed perturbation of the same mean
leaves it at the $\SI{2.94}{\angstrom}$ Nyquist limit; at a mean of $\SI{20}{\degree}$ the two differ by more than $\SI{11}{\angstrom}$ (Figs.~\ref{fig:bench2}C and \ref{fig:shape}). The pose estimators can tolerate a few catastrophically wrong orientations and average away but uniform errors on every particle distort reconstruction. cryoPARES carries the largest mean angular error measured here, $\SI{41.3}{\degree}$, and still produces good maps at $\SI{3.48}{\angstrom}$, better than a uniform $\SI{10}{\degree}$ blur, while \methodname{} and the classical filter sit on top of one another at $\SI{22}{\degree}$ and $\SI{3.63}{\angstrom}$. 


\paragraph{Restoration as a downstream operator}
Here we test whether \methodname{} is compatible with a learned 3D cryo-EM restorer, which benefits downstream refinement. EMReady \cite{he2023emready} raises the shell-correlation area from $0.674$ to $0.697$ and the masked correlation to $0.809$ (Fig.~\ref{fig:restore}), and cryoFM \cite{zhou2025cryofm} sharpens the grid-argmax reconstruction, before refinement, from $\SI{4.10}{\angstrom}$ to $\SI{3.68}{\angstrom}$. On strongly shuffled poses, both restorers suffer on the resulting low-quality reconstructions. Furthermore, the two restorers do not behave the same way when combined with different pose estimators (Fig.~\ref{fig:maps}C): EMReady moves three separate estimated maps in nearly the same direction, mean cosine $0.64$, while cryoFM's displacements are near-orthogonal at $0.06$. On the pose-estimator's manifold (supplementary Fig.~\ref{fig:smethod}), maps from different pose estimators fall close together compared to the distorted maps. 

Overall, \methodname{} works straightforward from 2D to 3D -- particles pose-estimation and back-projected, while a restorer works merely from 3D to 3D. By processing directly from the data source rather than from a prior over densities, our model potentially eliminates restorer-based biases for final restoration.

\subsection{Conformational heterogeneity signal}\label{sec:res:hetero}
We also show the capability of \methodname{} on treating conformational heterogeneity problem. As cryoEM structures are flexible molecules in nature, a pose estimator should capture a diverse population of the poses. We ran RECOVAR \cite{gilles2025recovar} twice on identical particles with identical translations while varying the rotations, and compared the two per-particle embeddings by canonical correlation \cite{hotelling1936relations}. Canonical correlation pairs each axis of one embedding with the axis of the other it best matches across particles, and ranks the pairs by how well they agree. Each pair is one collective mode of motion, and the ranking is blind to the arbitrary linear map a latent space is defined up to.

On EMPIAR-10076 the leading correlation between the embedding computed from our poses and the one computed from the deposited poses is $0.97$, and it stays consistent from two to twenty latent dimensions while the mean over all directions falls (Fig.~\ref{fig:bench2}A). What the declining mean measures is where extra canonical directions stop carrying conformational signal and start carrying noise, so the motion our poses preserve is concentrated in the few canonical directions.

On EMPIAR-10409, where our angular error is an order of magnitude larger, our poses retain $0.323$ of the embedding against $0.311$ for the classical matched filter and $0.248$ for cryoPARES, with disjoint bootstrap intervals and a floor of $0.02$ from the same pipeline run on the same particles with their poses shuffled (Fig.~\ref{fig:bench}B). Here, \methodname{} leads other estimators with comparable angular errors above; cryoPARES holds the best median angular error on this target (Sec.~\ref{sec:res:amortization}) but keeps the least conformational signal, $23\%$ less than our poses. Being accurate about one consensus structure does not guarantee the ability to capture the dynamics around it, and thus, \methodname{} is well-suited to generalize across specimens while preserving heterogeneity signal.

The conformational landscape survives as well as the individual poses.
Fig.~(\ref{fig:land}) plots RECOVAR's deconvolved density for the same particles embedded twice, with the deposited poses and with ours. Matching basins between the two pose sets by map correlation gives $0.92$--$0.99$, and their mobility fields agree at $0.90$--$0.94$. \methodname{}'s poses therefore reproduce not only the average structure but where the molecule is localized.

\subsection{Information scaling on orientation and transfer learning}\label{sec:res:theory}

Amortization across proteins is workable to the extent that something invariant is learnable. A natural candidate is the way orientation information is distributed over spatial frequency, which follows from the central-slice theorem and the microscope rather than from the molecule.

\paragraph{Orientation information grows as $k^2$.}
Eq.~\eqref{eq:exponent} predicts an exponent of exactly $2$ with no free parameter. Measured on real particles at their deposited orientations, across models that differ in input channels, training length and box size, the exponent is $1.89$ with an autocorrelation-robust standard error of $0.042$ (Fig.~\ref{fig:theory}A). The scaling that makes one estimator viable across proteins is obeyed by the data itself, with no network involved, letting an amortized model like ARCHER processing structures that were never trained on.

\paragraph{Where the remaining error lives.}
The error that remains is a selection problem. Refining the top candidates for each particle puts a pose within $\SI{15}{\degree}$ of the truth for $98\%$ of particles, while the matched-filter score picks that pose for only $92\%$ (Fig.~\ref{fig:theory}B). The right answer is usually already in the candidate set and the ranking rule discards it, so the improvement available to \methodname{} is in scoring rather than in a finer grid or a larger network.

\paragraph{Two angular scales.}
The score landscape around the true pose sets the two scales the pipeline works between (Fig.~\ref{fig:theory}C). The grid spacing of $7.4^\circ$ fixes what a discrete search can resolve, and the curvature of the score fixes what a continuous step can add below it. A derivative-free Newton step on three body axes carries the estimate from $\SI{4.5}{\degree}$ to $\SI{2.5}{\degree}$ on EMPIAR-10076, a $44\%$ reduction in median error and the largest single gain in the pipeline (supplementary Fig.~\ref{fig:s1}).

\paragraph{Sampling and measurement limits.}
Here, we look deeper into the two limits that are much closer across pose estimators than just comparing the pose error: the resolution the working box can support, and the precision with which shell correlation can tell two estimators apart. 

First, the pose estimate is done on an $L=64$ crop, and no pose can carry a reconstruction past that crop's Nyquist frequency. Across the five CESPED targets the reconstructions sit within a factor of $1.00$ to $1.39$ of it, and a consensus reconstruction from the deposited poses at the same box reaches the same limit, so further pose accuracy would not be meaningful (Fig.~\ref{fig:cross}). 

The second limit is the measurement itself (Fig.~\ref{fig:physics}). A fixed $0.143$ threshold holds every shell to a standard the innermost ones cannot meet, since a shell at radius $k$ holds only about $4\pi k^2$ samples, which is what the $\tfrac{1}{2}$-bit criterion corrects by scaling the threshold with that count \cite{vanheel2005fourier}, and a solvent mask (applied before the transform) raises the correlation while making it less certain -- where the inflation high-resolution noise substitution was introduced to detect \cite{chen2013highresolution}. Scored both ways on the same reconstruction the reported resolution moves from $\SI{3.57}{\angstrom}$ unmasked to $\SI{2.94}{\angstrom}$ masked, by at least $\SI{0.63}{\angstrom}$ since the masked curve never crosses $0.143$ within the band, while the three pose estimators scored through that same routine span $\SI{0.12}{\angstrom}$.\footnote{This routine and the benchmark's differ in shell binning, and the two put the same unmasked reconstruction at $\SI{3.57}{\angstrom}$ and $\SI{3.63}{\angstrom}$ respectively (Table~\ref{tab:maptargets}). Both comparisons above are made within one routine, so the offset cancels.} The choice of mask moves the headline number five times further than the choice of method does, which is the sense in which these estimators are not separated by resolution. Separating them needs a quantity beyond angular error, and retained conformational signal is the one this work finds discriminating.

\section{Discussion}\label{sec:discussion}

Conditioning a pose estimator on the reference volume moves the specimen out of the weights and into the input. The consequence is that one set of parameters serves proteins that were never in training, at accuracy competitive with methods fit to each dataset.

The argument has both theoretical and empirical support. Bayes' rule applied to Eq.~\eqref{eq:ell} makes $p(\Rot\mid y,\Vol)$ a single functional of the transfer function, the noise spectrum and the slice geometry, with the specimen entering multiplicatively through $P_k$ and $R_g$. The likelihood is not specimen-dependent, and a per-structure method obeys the same Eq.~\eqref{eq:k2law} that we do -- so the frequency scaling explains what a pose estimator must be sensitive to, not why amortizing across proteins should work.

We have presented empirical results on amortization. The first is that a finite-capacity network approximates that functional uniformly enough over the distribution of references to be useful on proteins outside its training set, while the second being that pooling specimens buys more in variance than it costs in the bias of sharing capacity between them. Neither follows from Eq.~\eqref{eq:k2law}.
Both are what the held-out and cross-target results measure, and they are also why a per-specimen estimator such as cryoPARES retains an advantage in angular error on the dataset it was fitted to. Thus, our shared network across specimen shows advantageous and transferable performance across targets.

This clarifies the relationship to existing families. Per-dataset amortized methods \cite{levy2022cryofire,zhong2021cryodrgn2} learn an encoder and a volume jointly, which is the right design when no reference exists and the goal is ab-initio structure. Supervised per-dataset estimators \cite{sanchezgarcia2025cryopares} learn the same matching operator we do but tie it to one molecule. \methodname{} occupies the regime where an initial reference is available, e.g., a consensus map, a homologue, or a predicted structure \cite{jumper2021alphafold,lin2023esmfold,proteinrediff}, and the task is to assign orientations at scale. The three regimes are complementary, and the present results suggest the shared component is larger than the protein-specific one.


Although per-structure models indistinguishably obtain high resolution in reconstruction (Table~\ref{tab:maptargets}), what separate them is how much conformational signal survives to the downstream analysis, and that is the measure on which our poses lead every estimator compared here -- $0.323$ of the embedding retained against $0.311$ and $0.248$, with the most angularly accurate estimator retaining the least. For flexible assemblies that is the quantity of interest, and it should be reported alongside resolution.

Two limitations bound the scope. The estimator requires a reference, and the measurements locate that requirement precisely: it is needed to initialize the loop rather than to sustain it. After the first pass, every subsequent round scores particles against a volume rebuilt from our own estimates, and the accuracy attained with that self-generated reference matches the one attained against an independent reference built from the opposite half-set ($\SI{4.54}{\degree}$ against $\SI{4.53}{\degree}$ at the grid stage, and $\SI{2.60}{\degree}$ against $\SI{2.54}{\degree}$ after refinement), holding to within $\SI{0.1}{\degree}$ over three further rounds. Starting the same loop from a deliberately crude reference behaves differently: accuracy falls as the iterations proceed, because a poor reference yields a poor reconstruction and the error compounds. The reference must therefore be good enough to place the first estimate in the right basin -- a consensus map, a homologue, or a predicted structure -- after which the pipeline supplies its own. And the residual error is a selection problem: a correct pose sits in the candidate set for $98\%$ of particles while the score picks it for $92\%$, so a better ranking rule is the clearest available gain. A learned selector over refined candidates is the natural next step, and the measured headroom quantifies what it can deliver.


\section{Conclusion}\label{sec:conclusion}




Orientation assignments (and other problems) in cryoEM have been treated as a per-dataset problem because the reference volume has always been carried in the parameters of whatever performs the assignment. In this work, we introduced \methodname{} as a general solution for this problem, where it can be applied for structures (seen or unseen by the model). 

Notably, the main advantage here is not accuracy but rather generality. \methodname{} can be applied to various structures, but its cost is visible. Specifically, when a target's sampling does not bind, we trail behind a specimen-trained estimator by about $\SI{2}{\angstrom}$. When sampling does bind, angular error stops discriminating among competent estimators, and the measurement that continues to separate them is how much conformational signal reaches the downstream analysis. We propose reporting that alongside resolution whenever flexible assemblies are the object of study.

Overall, the performance shown by \methodname{} is not trivial. It is able to reconstruct unseen structures, using conditionals obtained from reference structures, within competitive error margins -- a notable feature for a rather general technique with respect to the prior mentioned related works.

\begin{acknowledgments}
The authors declare no competing interests.
\end{acknowledgments}

\section{Code Availability}
Codes and data are deposited under https://github.com/ndnng/ARCHER

\section*{References}
\bibliography{references}

@article{derosier1968reconstruction,
  title   = {{Reconstruction of three dimensional structures from electron micrographs}},
  author  = {De Rosier, D. J. and Klug, A.},
  journal = {Nature},
  volume  = {217},
  number  = {5124},
  pages   = {130--134},
  year    = {1968},
  doi     = {10.1038/217130a0}
}

@article{crowther1970reconstruction,
  title   = {{The reconstruction of a three-dimensional structure from projections and its application to electron microscopy}},
  author  = {Crowther, R. A. and DeRosier, D. J. and Klug, A.},
  journal = {Proceedings of the Royal Society of London A},
  volume  = {317},
  number  = {1530},
  pages   = {319--340},
  year    = {1970},
  doi     = {10.1098/rspa.1970.0119}
}

@book{frank2006three,
  title     = {{Three-Dimensional Electron Microscopy of Macromolecular Assemblies}},
  author    = {Frank, Joachim},
  edition   = {2},
  publisher = {Oxford University Press},
  address   = {New York},
  year      = {2006}
}

@article{penczek1994ribosome,
  title   = {{The ribosome at improved resolution: new techniques for merging and orientation refinement in {3D} cryo-electron microscopy of biological particles}},
  author  = {Penczek, Pawel A. and Grassucci, Robert A. and Frank, Joachim},
  journal = {Ultramicroscopy},
  volume  = {53},
  number  = {3},
  pages   = {251--270},
  year    = {1994},
  doi     = {10.1016/0304-3991(94)90038-8}
}

@article{sigworth1998maximum,
  title   = {{A maximum-likelihood approach to single-particle image refinement}},
  author  = {Sigworth, F. J.},
  journal = {Journal of Structural Biology},
  volume  = {122},
  number  = {3},
  pages   = {328--339},
  year    = {1998},
  doi     = {10.1006/jsbi.1998.4014}
}

@article{scheres2012relion,
  title   = {{RELION}: {Implementation of a Bayesian approach to cryo-EM structure determination}},
  author  = {Scheres, Sjors H. W.},
  journal = {Journal of Structural Biology},
  volume  = {180},
  number  = {3},
  pages   = {519--530},
  year    = {2012},
  doi     = {10.1016/j.jsb.2012.09.006}
}

@article{scheres2012bayesian,
  title   = {{A Bayesian view on cryo-EM structure determination}},
  author  = {Scheres, Sjors H. W.},
  journal = {Journal of Molecular Biology},
  volume  = {415},
  number  = {2},
  pages   = {406--418},
  year    = {2012},
  doi     = {10.1016/j.jmb.2011.11.010}
}

@article{punjani2017cryosparc,
  title   = {{cryoSPARC}: {algorithms for rapid unsupervised cryo-EM structure determination}},
  author  = {Punjani, Ali and Rubinstein, John L. and Fleet, David J. and Brubaker, Marcus A.},
  journal = {Nature Methods},
  volume  = {14},
  number  = {3},
  pages   = {290--296},
  year    = {2017},
  doi     = {10.1038/nmeth.4169}
}

@article{grigorieff2007frealign,
  title   = {{FREALIGN}: {high-resolution refinement of single particle structures}},
  author  = {Grigorieff, Nikolaus},
  journal = {Journal of Structural Biology},
  volume  = {157},
  number  = {1},
  pages   = {117--125},
  year    = {2007},
  doi     = {10.1016/j.jsb.2006.05.004}
}

@article{rohou2015ctffind4,
  title   = {{CTFFIND4}: {Fast and accurate defocus estimation from electron micrographs}},
  author  = {Rohou, Alexis and Grigorieff, Nikolaus},
  journal = {Journal of Structural Biology},
  volume  = {192},
  number  = {2},
  pages   = {216--221},
  year    = {2015},
  doi     = {10.1016/j.jsb.2015.08.008}
}

@article{rosenthal2003optimal,
  title   = {{Optimal determination of particle orientation, absolute hand, and contrast loss in single-particle electron cryomicroscopy}},
  author  = {Rosenthal, Peter B. and Henderson, Richard},
  journal = {Journal of Molecular Biology},
  volume  = {333},
  number  = {4},
  pages   = {721--745},
  year    = {2003},
  doi     = {10.1016/j.jmb.2003.07.013}
}

@article{vanheel2005fourier,
  title   = {{Fourier shell correlation threshold criteria}},
  author  = {van Heel, Marin and Schatz, Michael},
  journal = {Journal of Structural Biology},
  volume  = {151},
  number  = {3},
  pages   = {250--262},
  year    = {2005},
  doi     = {10.1016/j.jsb.2005.05.009}
}

@article{kimanius2024blush,
  title   = {{Data-driven regularization lowers the size barrier of cryo-EM structure determination}},
  author  = {Kimanius, Dari and Jamali, Kiarash and Wilkinson, Max E. and L{\"o}vestam, Sofia and Velazhahan, Vaithish and Nakane, Takanori and Scheres, Sjors H. W.},
  journal = {Nature Methods},
  volume  = {21},
  pages   = {1216--1221},
  year    = {2024},
  doi     = {10.1038/s41592-024-02304-8}
}

@article{derosier2000correction,
  title   = {{Correction of high-resolution data for curvature of the Ewald sphere}},
  author  = {DeRosier, D. J.},
  journal = {Ultramicroscopy},
  volume  = {81},
  number  = {2},
  pages   = {83--98},
  year    = {2000},
  doi     = {10.1016/S0304-3991(99)00120-5}
}

@article{wolf2006ewald,
  title   = {{Ewald sphere correction for single-particle electron microscopy}},
  author  = {Wolf, Matthias and DeRosier, David J. and Grigorieff, Nikolaus},
  journal = {Ultramicroscopy},
  volume  = {106},
  number  = {4-5},
  pages   = {376--382},
  year    = {2006},
  doi     = {10.1016/j.ultramic.2005.11.001}
}

@article{garciacondado2022handedness,
  title   = {{Automatic determination of the handedness of single-particle maps of macromolecules solved by CryoEM}},
  author  = {Garcia Condado, J. and Mu{\~n}oz-Barrutia, A. and Sorzano, C. O. S.},
  journal = {Journal of Structural Biology},
  volume  = {214},
  number  = {4},
  pages   = {107915},
  year    = {2022},
  doi     = {10.1016/j.jsb.2022.107915}
}

@article{singer2011three,
  title   = {{Three-dimensional structure determination from common lines in cryo-EM by eigenvectors and semidefinite programming}},
  author  = {Singer, A. and Shkolnisky, Y.},
  journal = {SIAM Journal on Imaging Sciences},
  volume  = {4},
  number  = {2},
  pages   = {543--572},
  year    = {2011},
  doi     = {10.1137/090767777}
}

@article{wang2013orientation,
  title   = {{Orientation determination of cryo-EM images using least unsquared deviations}},
  author  = {Wang, Lanhui and Singer, Amit and Wen, Zaiwen},
  journal = {SIAM Journal on Imaging Sciences},
  volume  = {6},
  number  = {4},
  pages   = {2450--2483},
  year    = {2013},
  doi     = {10.1137/130916436}
}

@article{wang2013firm,
  title   = {{A Fourier-based approach for iterative 3D reconstruction from cryo-EM images}},
  author  = {Wang, Lanhui and Shkolnisky, Yoel and Singer, Amit},
  journal = {arXiv preprint arXiv:1307.5824},
  year    = {2013}
}

@article{gilles2025recovar,
  title   = {{Cryo-EM heterogeneity analysis using regularized covariance estimation and kernel regression}},
  author  = {Gilles, Marc Aur{\`e}le and Singer, Amit},
  journal = {Proceedings of the National Academy of Sciences},
  volume  = {122},
  number  = {9},
  pages   = {e2419140122},
  year    = {2025},
  doi     = {10.1073/pnas.2419140122}
}

@article{zhong2021cryodrgn,
  title   = {{CryoDRGN}: {reconstruction of heterogeneous cryo-EM structures using neural networks}},
  author  = {Zhong, Ellen D. and Bepler, Tristan and Berger, Bonnie and Davis, Joseph H.},
  journal = {Nature Methods},
  volume  = {18},
  number  = {2},
  pages   = {176--185},
  year    = {2021},
  doi     = {10.1038/s41592-020-01049-4}
}

@inproceedings{zhong2021cryodrgn2,
  title     = {{CryoDRGN2}: {Ab initio} neural reconstruction of {3D} protein structures from real cryo-{EM} images},
  author    = {Zhong, Ellen D. and Lerer, Adam and Davis, Joseph H. and Berger, Bonnie},
  booktitle = {Proceedings of the IEEE/CVF International Conference on Computer Vision (ICCV)},
  pages     = {4066--4075},
  year      = {2021}
}

@inproceedings{levy2022cryoai,
  title     = {{CryoAI}: {Amortized} inference of poses for ab initio reconstruction of {3D} molecular volumes from real cryo-{EM} images},
  author    = {Levy, Axel and Poitevin, Fr{\'e}d{\'e}ric and Martel, Julien and Nashed, Youssef and Peck, Ariana and Miolane, Nina and Ratner, Daniel and Dunne, Mike and Wetzstein, Gordon},
  booktitle = {European Conference on Computer Vision (ECCV)},
  pages     = {540--557},
  year      = {2022}
}

@inproceedings{levy2022cryofire,
  title     = {{Amortized inference for heterogeneous reconstruction in cryo-EM}},
  author    = {Levy, Axel and Wetzstein, Gordon and Martel, Julien and Poitevin, Frederic and Zhong, Ellen D.},
  booktitle = {Advances in Neural Information Processing Systems (NeurIPS)},
  year      = {2022}
}

@inproceedings{shekarforoush2024cryospin,
  title     = {{CryoSPIN}: {Improving} ab-initio cryo-{EM} reconstruction with semi-amortized pose inference},
  author    = {Shekarforoush, Shayan and Lindell, David B. and Brubaker, Marcus A. and Fleet, David J.},
  booktitle = {Advances in Neural Information Processing Systems (NeurIPS)},
  year      = {2024}
}

@article{sanchezgarcia2023cesped,
  title   = {{CESPED}: {a} new benchmark for supervised particle pose estimation in cryo-{EM}},
  author  = {Sanchez-Garcia, Ruben and Marsden, Joel and Stagg, Scott M. and Waterman, David and Cowtan, Kevin and Deane, Charlotte M.},
  journal = {arXiv preprint arXiv:2311.06194},
  year    = {2023}
}

@article{sanchezgarcia2025cryopares,
  title   = {{Supervised deep learning for efficient cryo-EM image alignment in drug discovery}},
  author  = {Sanchez-Garcia, Ruben and Deane, Charlotte M.},
  journal = {bioRxiv},
  year    = {2025},
  doi     = {10.1101/2025.03.04.641536}
}

@article{banjac2021learning,
  title   = {{Learning to recover orientations from projections in single-particle cryo-EM}},
  author  = {Banjac, Jelena and Donati, Laur{\`e}ne and Defferrard, Micha{\"e}l},
  journal = {arXiv preprint arXiv:2104.06237},
  year    = {2021}
}

@article{lian2022end,
  title   = {{End-to-end orientation estimation from 2D cryo-EM images}},
  author  = {Lian, Ruyi and Huang, Bingyao and Wang, Liguo and Liu, Qun and Lin, Yuewei and Ling, Haibin},
  journal = {Acta Crystallographica Section D},
  volume  = {78},
  number  = {2},
  pages   = {174--186},
  year    = {2022},
  doi     = {10.1107/S2059798321011761}
}

@inproceedings{zhang2025cryofastar,
  title     = {{CryoFastAR}: {Fast} cryo-{EM} ab initio reconstruction made easy},
  author    = {Zhang, Jiakai and Zhou, Shouchen and Dai, Haizhao and Liu, Xinhang and Wang, Peihao and Fan, Zhiwen and Pei, Yuan and Yu, Jingyi},
  booktitle = {Proceedings of the IEEE/CVF International Conference on Computer Vision (ICCV)},
  year      = {2025}
}

@article{yan2026cryoief,
  title   = {{A comprehensive foundation model for cryo-EM image processing}},
  author  = {Yan, Yang and Fan, Shiqi and Yuan, Fajie and Shen, Huaizong},
  journal = {Nature Methods},
  volume  = {23},
  pages   = {88--95},
  year    = {2026},
  doi     = {10.1038/s41592-025-02916-8}
}

@article{zhou2025cryofm,
  title   = {{CryoFM}: {A} flow-based foundation model for cryo-{EM} densities},
  author  = {Zhou, Yi and Li, Yilai and Yuan, Jing and Gu, Quanquan},
  journal = {arXiv preprint arXiv:2409.09992},
  year    = {2025}
}

@article{qu2025cryonerf,
  title   = {{CryoNeRF}: {reconstruction} of homogeneous and heterogeneous cryo-{EM} structures using neural radiance field},
  author  = {Qu, Huaizhi and Wang, Xiao and Zhang, Yuanyuan and Wang, Sheng and Noble, William Stafford and Chen, Tianlong},
  journal = {bioRxiv},
  year    = {2025},
  doi     = {10.1101/2025.01.10.632460}
}

@inproceedings{huang2026cryonetrefine,
  title     = {{CryoNet.Refine}: {A} one-step diffusion model for rapid refinement of structural models with cryo-{EM} density map restraints},
  author    = {Huang, Fuyao and Yu, Xiaozhu and Xu, Kui and Zhang, Qiangfeng Cliff},
  booktitle = {International Conference on Learning Representations (ICLR)},
  year      = {2026}
}

@article{li2024cryoddm,
  title   = {{CryoDDM}: {CryoEM} denoising diffusion model for heterogeneous conformational reconstruction},
  author  = {Li, Fuwei and Chen, Yuanbo and Dong, Hao and Ji, Chenxuan and Wang, Xinsheng and Zhang, Chuanyang and Wang, Zupeng and Hu, Bin and Zhang, Fa and Wan, Xiaohua},
  journal = {arXiv preprint},
  year    = {2024}
}

@article{gorski2005healpix,
  title   = {{HEALPix}: {A} framework for high-resolution discretization and fast analysis of data distributed on the sphere},
  author  = {G{\'o}rski, K. M. and Hivon, E. and Banday, A. J. and Wandelt, B. D. and Hansen, F. K. and Reinecke, M. and Bartelmann, M.},
  journal = {The Astrophysical Journal},
  volume  = {622},
  number  = {2},
  pages   = {759--771},
  year    = {2005},
  doi     = {10.1086/427976}
}

@article{yershova2010generating,
  title   = {{Generating uniform incremental grids on SO(3) using the Hopf fibration}},
  author  = {Yershova, Anna and Jain, Swati and LaValle, Steven M. and Mitchell, Julie C.},
  journal = {The International Journal of Robotics Research},
  volume  = {29},
  number  = {7},
  pages   = {801--812},
  year    = {2010},
  doi     = {10.1177/0278364909352700}
}

@inproceedings{murphy2021implicit,
  title     = {{Implicit-PDF}: {Non-parametric} representation of probability distributions on the rotation manifold},
  author    = {Murphy, Kieran A. and Esteves, Carlos and Jampani, Varun and Ramalingam, Srikumar and Makadia, Ameesh},
  booktitle = {International Conference on Machine Learning (ICML)},
  year      = {2021}
}

@inproceedings{klee2023image2sphere,
  title     = {{Image to Sphere}: {Learning} equivariant features for efficient pose prediction},
  author    = {Klee, David M. and Biza, Ondrej and Platt, Robert and Walters, Robin},
  booktitle = {International Conference on Learning Representations (ICLR)},
  year      = {2023}
}

@inproceedings{he2016deep,
  title     = {{Deep residual learning for image recognition}},
  author    = {He, Kaiming and Zhang, Xiangyu and Ren, Shaoqing and Sun, Jian},
  booktitle = {Proceedings of the IEEE Conference on Computer Vision and Pattern Recognition (CVPR)},
  pages     = {770--778},
  year      = {2016}
}

@inproceedings{wu2018group,
  title     = {{Group normalization}},
  author    = {Wu, Yuxin and He, Kaiming},
  booktitle = {European Conference on Computer Vision (ECCV)},
  pages     = {3--19},
  year      = {2018}
}

@article{oord2018representation,
  title   = {{Representation learning with contrastive predictive coding}},
  author  = {van den Oord, Aaron and Li, Yazhe and Vinyals, Oriol},
  journal = {arXiv preprint arXiv:1807.03748},
  year    = {2018}
}

@inproceedings{radford2021learning,
  title     = {{Learning transferable visual models from natural language supervision}},
  author    = {Radford, Alec and Kim, Jong Wook and Hallacy, Chris and Ramesh, Aditya and Goh, Gabriel and Agarwal, Sandhini and Sastry, Girish and Askell, Amanda and Mishkin, Pamela and Clark, Jack and Krueger, Gretchen and Sutskever, Ilya},
  booktitle = {International Conference on Machine Learning (ICML)},
  pages     = {8748--8763},
  year      = {2021}
}

@inproceedings{loshchilov2017sgdr,
  title     = {{SGDR}: {Stochastic} gradient descent with warm restarts},
  author    = {Loshchilov, Ilya and Hutter, Frank},
  booktitle = {International Conference on Learning Representations (ICLR)},
  year      = {2017}
}

@inproceedings{loshchilov2019decoupled,
  title     = {{Decoupled weight decay regularization}},
  author    = {Loshchilov, Ilya and Hutter, Frank},
  booktitle = {International Conference on Learning Representations (ICLR)},
  year      = {2019}
}

@inproceedings{szegedy2016rethinking,
  title     = {{Rethinking the inception architecture for computer vision}},
  author    = {Szegedy, Christian and Vanhoucke, Vincent and Ioffe, Sergey and Shlens, Jon and Wojna, Zbigniew},
  booktitle = {Proceedings of the IEEE Conference on Computer Vision and Pattern Recognition (CVPR)},
  pages     = {2818--2826},
  year      = {2016}
}

@article{dempster1977maximum,
  title   = {{Maximum likelihood from incomplete data via the EM algorithm}},
  author  = {Dempster, A. P. and Laird, N. M. and Rubin, D. B.},
  journal = {Journal of the Royal Statistical Society: Series B},
  volume  = {39},
  number  = {1},
  pages   = {1--22},
  year    = {1977}
}

@article{iudin2016empiar,
  title   = {{EMPIAR}: {a} public archive for raw electron microscopy image data},
  author  = {Iudin, Andrii and Korir, Paul K. and Salavert-Torres, Jos{\'e} and Kleywegt, Gerard J. and Patwardhan, Ardan},
  journal = {Nature Methods},
  volume  = {13},
  number  = {5},
  pages   = {387--388},
  year    = {2016},
  doi     = {10.1038/nmeth.3806}
}

@article{lawson2016emdatabank,
  title   = {{EMDataBank unified data resource for 3DEM}},
  author  = {Lawson, Catherine L. and Patwardhan, Ardan and Baker, Matthew L. and Hryc, Corey and Garcia, Eduardo Sanz and Hudson, Brian P. and Lagerstedt, Ingvar and Ludtke, Steven J. and Pintilie, Grigore and Sala, Raul and Westbrook, John D. and Berman, Helen M. and Kleywegt, Gerard J. and Chiu, Wah},
  journal = {Nucleic Acids Research},
  volume  = {44},
  number  = {D1},
  pages   = {D396--D403},
  year    = {2016},
  doi     = {10.1093/nar/gkv1126}
}

@article{chen2013highresolution,
  title   = {High-resolution noise substitution to measure overfitting and validate resolution in 3{D} structure determination by single particle electron cryomicroscopy},
  author  = {Chen, Shaoxia and McMullan, Greg and Faruqi, Abdul R. and Murshudov, Garib N. and Short, Judith M. and Scheres, Sjors H. W. and Henderson, Richard},
  journal = {Ultramicroscopy},
  volume  = {135},
  pages   = {24--35},
  year    = {2013},
  doi     = {10.1016/j.ultramic.2013.06.004}
}

@incollection{scheres2010classification,
  title     = {Classification of structural heterogeneity by maximum-likelihood methods},
  author    = {Scheres, Sjors H. W.},
  booktitle = {Methods in Enzymology},
  volume    = {482},
  pages     = {295--320},
  year      = {2010},
  publisher = {Elsevier},
  doi       = {10.1016/S0076-6879(10)82012-9}
}

@article{lyumkis2013likelihood,
  title   = {Likelihood-based classification of cryo-{EM} images using {FREALIGN}},
  author  = {Lyumkis, Dmitry and Brilot, Axel F. and Theobald, Douglas L. and Grigorieff, Nikolaus},
  journal = {Journal of Structural Biology},
  volume  = {183},
  number  = {3},
  pages   = {377--388},
  year    = {2013},
  doi     = {10.1016/j.jsb.2013.07.005}
}

@article{henderson2012outcome,
  title   = {Outcome of the first electron microscopy validation task force meeting},
  author  = {Henderson, Richard and Sali, Andrej and Baker, Matthew L. and others},
  journal = {Structure},
  volume  = {20},
  number  = {2},
  pages   = {205--214},
  year    = {2012},
  doi     = {10.1016/j.str.2011.12.014}
}

@article{he2023emready,
  title   = {Improvement of cryo-{EM} maps by simultaneous local and non-local deep learning},
  author  = {He, Jiahua and Li, Tao and Huang, Sheng-You},
  journal = {Nature Communications},
  volume  = {14},
  number  = {1},
  pages   = {3217},
  year    = {2023},
  doi     = {10.1038/s41467-023-39031-1}
}

@article{sanchezgarcia2021deepemhancer,
  title   = {{DeepEMhancer}: a deep learning solution for cryo-{EM} volume post-processing},
  author  = {Sanchez-Garcia, Ruben and Gomez-Blanco, Josue and Cuervo, Ana and
             Carazo, Jose Maria and Sorzano, Carlos Oscar S. and Vargas, Javier},
  journal = {Communications Biology},
  volume  = {4},
  number  = {1},
  pages   = {874},
  year    = {2021},
  doi     = {10.1038/s42003-021-02399-1}
}

@article{kimanius2021relion4,
  title   = {New tools for automated cryo-{EM} single-particle analysis in {RELION-4.0}},
  author  = {Kimanius, Dari and Dong, Liyi and Sharov, Grigory and Nakane, Takanori and
             Scheres, Sjors H. W.},
  journal = {Biochemical Journal},
  volume  = {478},
  number  = {24},
  pages   = {4169--4185},
  year    = {2021},
  doi     = {10.1042/BCJ20210708}
}

@article{jumper2021alphafold,
  title   = {Highly accurate protein structure prediction with {AlphaFold}},
  author  = {Jumper, John and Evans, Richard and Pritzel, Alexander and Green, Tim and
             Figurnov, Michael and Ronneberger, Olaf and Tunyasuvunakool, Kathryn and
             Bates, Russ and {\v{Z}}{\'i}dek, Augustin and Potapenko, Anna and others},
  journal = {Nature},
  volume  = {596},
  number  = {7873},
  pages   = {583--589},
  year    = {2021},
  doi     = {10.1038/s41586-021-03819-2}
}

@article{lin2023esmfold,
  title   = {Evolutionary-scale prediction of atomic-level protein structure with a language model},
  author  = {Lin, Zeming and Akin, Halil and Rao, Roshan and Hie, Brian and Zhu, Zhongkai and
             Lu, Wenting and Smetanin, Nikita and Verkuil, Robert and Kabeli, Ori and
             Shmueli, Yaniv and others},
  journal = {Science},
  volume  = {379},
  number  = {6637},
  pages   = {1123--1130},
  year    = {2023},
  doi     = {10.1126/science.ade2574}
}

@inproceedings{ronneberger2015unet,
  title     = {U-Net: Convolutional networks for biomedical image segmentation},
  author    = {Ronneberger, Olaf and Fischer, Philipp and Brox, Thomas},
  booktitle = {Medical Image Computing and Computer-Assisted Intervention (MICCAI)},
  pages     = {234--241},
  year      = {2015}
}

@inproceedings{ho2020ddpm,
  title     = {Denoising diffusion probabilistic models},
  author    = {Ho, Jonathan and Jain, Ajay and Abbeel, Pieter},
  booktitle = {Advances in Neural Information Processing Systems (NeurIPS)},
  volume    = {33},
  pages     = {6840--6851},
  year      = {2020}
}

@inproceedings{nichol2021improved,
  title     = {Improved denoising diffusion probabilistic models},
  author    = {Nichol, Alexander Quinn and Dhariwal, Prafulla},
  booktitle = {International Conference on Machine Learning (ICML)},
  pages     = {8162--8171},
  year      = {2021}
}

@article{wiener1949extrapolation,
  title     = {Extrapolation, Interpolation, and Smoothing of Stationary Time Series},
  author    = {Wiener, Norbert},
  journal   = {MIT Press},
  year      = {1949}
}

@article{hotelling1936relations,
  title   = {Relations between two sets of variates},
  author  = {Hotelling, Harold},
  journal = {Biometrika},
  volume  = {28},
  number  = {3--4},
  pages   = {321--377},
  year    = {1936}
}

@article{scheres2012goldstandard,
  title   = {Prevention of overfitting in cryo-{EM} structure determination},
  author  = {Scheres, Sjors H. W. and Chen, Shaoxia},
  journal = {Nature Methods},
  volume  = {9},
  number  = {9},
  pages   = {853--854},
  year    = {2012},
  doi     = {10.1038/nmeth.2115}
}

@inproceedings{kingma2014vae,
  title     = {Auto-encoding variational {B}ayes},
  author    = {Kingma, Diederik P. and Welling, Max},
  booktitle = {International Conference on Learning Representations (ICLR)},
  year      = {2014}
}

@inproceedings{cremer2018amortization,
  title     = {Inference suboptimality in variational autoencoders},
  author    = {Cremer, Chris and Li, Xuechen and Duvenaud, David},
  booktitle = {International Conference on Machine Learning (ICML)},
  pages     = {1078--1086},
  year      = {2018}
}

@article{punjani2023threedflex,
  title   = {{3DFlex}: determining structure and motion of flexible proteins from cryo-{EM}},
  author  = {Punjani, Ali and Fleet, David J.},
  journal = {Nature Methods},
  volume  = {20},
  number  = {6},
  pages   = {860--870},
  year    = {2023},
  doi     = {10.1038/s41592-023-01853-8}
}

@article{nakane2018multibody,
  title   = {Characterisation of molecular motions in cryo-{EM} single-particle data by
             multi-body refinement in {RELION}},
  author  = {Nakane, Takanori and Kimanius, Dari and Lindahl, Erik and Scheres, Sjors H. W.},
  journal = {eLife},
  volume  = {7},
  pages   = {e36861},
  year    = {2018},
  doi     = {10.7554/eLife.36861}
}

@article{punjani2020nonuniform,
  title   = {Non-uniform refinement: adaptive regularization improves single-particle cryo-{EM}
             reconstruction},
  author  = {Punjani, Ali and Zhang, Haowei and Fleet, David J.},
  journal = {Nature Methods},
  volume  = {17},
  number  = {12},
  pages   = {1214--1221},
  year    = {2020},
  doi     = {10.1038/s41592-020-00990-8}
}

@inproceedings{cohen2018spherical,
  title     = {Spherical {CNN}s},
  author    = {Cohen, Taco S. and Geiger, Mario and K{\"o}hler, Jonas and Welling, Max},
  booktitle = {International Conference on Learning Representations (ICLR)},
  year      = {2018}
}

@article{bepler2019topaz,
  title   = {Positive-unlabeled convolutional neural networks for particle picking in
             cryo-electron micrographs},
  author  = {Bepler, Tristan and Morin, Andrew and Rapp, Micah and Brasch, Julia and
             Shapiro, Lawrence and Noble, Alex J. and Berger, Bonnie},
  journal = {Nature Methods},
  volume  = {16},
  number  = {11},
  pages   = {1153--1160},
  year    = {2019},
  doi     = {10.1038/s41592-019-0575-8}
}

@article{tegunov2019warp,
  title   = {Real-time cryo-electron microscopy data preprocessing with {Warp}},
  author  = {Tegunov, Dimitry and Cramer, Patrick},
  journal = {Nature Methods},
  volume  = {16},
  number  = {11},
  pages   = {1146--1152},
  year    = {2019},
  doi     = {10.1038/s41592-019-0580-y}
}

@article{rao1945information,
  title   = {Information and the accuracy attainable in the estimation of statistical parameters},
  author  = {Rao, C. Radhakrishna},
  journal = {Bulletin of the Calcutta Mathematical Society},
  volume  = {37},
  pages   = {81--91},
  year    = {1945}
}

@book{cramer1946mathematical,
  title     = {Mathematical Methods of Statistics},
  author    = {Cram{\'e}r, Harald},
  publisher = {Princeton University Press},
  year      = {1946}
}

@article{becker1997topology,
  title   = {The topology of multidimensional potential energy surfaces: Theory and application to peptide structure and kinetics},
  author  = {Becker, Oren M. and Karplus, Martin},
  journal = {The Journal of Chemical Physics},
  volume  = {106},
  number  = {4},
  pages   = {1495--1517},
  year    = {1997}
}

@article{kramers1940brownian,
  title   = {Brownian motion in a field of force and the diffusion model of chemical reactions},
  author  = {Kramers, H. A.},
  journal = {Physica},
  volume  = {7},
  number  = {4},
  pages   = {284--304},
  year    = {1940}
}

@article{bendory2020cryoem,
  title   = {{Single-particle cryo-electron microscopy: Mathematical theory, computational challenges, and opportunities}},
  author  = {Bendory, T. and Bartesaghi, A. and Singer, A.},
  journal = {IEEE Signal Processing Magazine},
  volume  = {37},
  number  = {2},
  pages   = {58--76},
  year    = {2020},
  doi     = {10.1109/MSP.2019.2957822}
}

@article{singer2020computational,
  title   = {{Computational methods for single-particle electron cryomicroscopy}},
  author  = {Singer, A. and Sigworth, F. J.},
  journal = {Annual Review of Biomedical Data Science},
  volume  = {3},
  pages   = {163--190},
  year    = {2020},
  doi     = {10.1146/annurev-biodatasci-021020-093826}
}

@article{kuhlbrandt2014resolution,
  title   = {The resolution revolution},
  author  = {K{\"u}hlbrandt, Werner},
  journal = {Science},
  volume  = {343},
  number  = {6178},
  pages   = {1443--1444},
  year    = {2014},
  doi     = {10.1126/science.1251652}
}

@article{nakane2020single,
  title   = {Single-particle cryo-{EM} at atomic resolution},
  author  = {Nakane, Takanori and Kotecha, Abhay and Sente, Andrija and McMullan, Greg and Masiulis, Simonas and Brown, Patricia M. G. E. and Grigoras, Ioana T. and Malinauskaite, Lina and Malinauskas, Tomas and Miehling, Jonas and Uchanski, Tomasz and Yu, Lingbo and Karia, Dimple and Pechnikova, Evgeniya V. and de Jong, Erwin and Keizer, Jeroen and Bischoff, Maarten and McCormack, Jamie and Tiemeijer, Peter and Hardwick, Steven W. and Chirgadze, Dimitri Y. and Murshudov, Garib and Aricescu, A. Radu and Scheres, Sjors H. W.},
  journal = {Nature},
  volume  = {587},
  number  = {7832},
  pages   = {152--156},
  year    = {2020},
  doi     = {10.1038/s41586-020-2829-0}
}

@article{yip2020atomic,
  title   = {Atomic-resolution protein structure determination by cryo-{EM}},
  author  = {Yip, Ka Man and Fischer, Niels and Paknia, Elham and Chari, Ashwin and Stark, Holger},
  journal = {Nature},
  volume  = {587},
  number  = {7832},
  pages   = {157--161},
  year    = {2020},
  doi     = {10.1038/s41586-020-2833-4}
}

@inproceedings{esteves2018learning,
  title     = {Learning {SO(3)} equivariant representations with spherical {CNN}s},
  author    = {Esteves, Carlos and Allen-Blanchette, Christine and Makadia, Ameesh and Daniilidis, Kostas},
  booktitle = {European Conference on Computer Vision (ECCV)},
  pages     = {52--68},
  year      = {2018},
  doi       = {10.1007/978-3-030-01261-8_4}
}

@inproceedings{bansal2023cold,
  title     = {Cold diffusion: inverting arbitrary image transforms without noise},
  author    = {Bansal, Arpit and Borgnia, Eitan and Chu, Hong-Min and Li, Jie S. and Kazemi, Hamid and Huang, Furong and Goldblum, Micah and Geiping, Jonas and Goldstein, Tom},
  booktitle = {Advances in Neural Information Processing Systems (NeurIPS)},
  year      = {2023}
}

@article{daras2023soft,
  title   = {Soft diffusion: score matching with general corruptions},
  author  = {Daras, Giannis and Delbracio, Mauricio and Talebi, Hossein and Dimakis, Alexandros G. and Milanfar, Peyman},
  journal = {Transactions on Machine Learning Research},
  year    = {2023}
}

@article{delbracio2023inversion,
  title   = {Inversion by direct iteration: an alternative to denoising diffusion for image restoration},
  author  = {Delbracio, Mauricio and Milanfar, Peyman},
  journal = {Transactions on Machine Learning Research},
  year    = {2023}
}

@inproceedings{liu2023i2sb,
  title     = {{I$^2$SB}: image-to-image {Schr{\"o}dinger} bridge},
  author    = {Liu, Guan-Horng and Vahdat, Arash and Huang, De-An and Theodorou, Evangelos A. and Nie, Weili and Anandkumar, Anima},
  booktitle = {International Conference on Machine Learning (ICML)},
  year      = {2023}
}

@inproceedings{luo2023image,
  title     = {Image restoration with mean-reverting stochastic differential equations},
  author    = {Luo, Ziwei and Gustafsson, Fredrik K. and Zhao, Zheng and Sj{\"o}lund, Jens and Sch{\"o}n, Thomas B.},
  booktitle = {International Conference on Machine Learning (ICML)},
  year      = {2023}
}

@article{rickgauer2017single,
  title   = {Single-protein detection in crowded molecular environments in cryo-{EM} images},
  author  = {Rickgauer, J. Peter and Grigorieff, Nikolaus and Denk, Winfried},
  journal = {eLife},
  volume  = {6},
  pages   = {e25648},
  year    = {2017},
  doi     = {10.7554/eLife.25648}
}

@article{lucas2021locating,
  title   = {Locating macromolecular assemblies in cells by 2{D} template matching with {cisTEM}},
  author  = {Lucas, Bronwyn A. and Himes, Benjamin A. and Xue, Liang and Grant, Timothy and Mahamid, Julia and Grigorieff, Nikolaus},
  journal = {eLife},
  volume  = {10},
  pages   = {e68946},
  year    = {2021},
  doi     = {10.7554/eLife.68946}
}

@article{baldwin2020non,
  title   = {Non-uniformity of projection distributions attenuates resolution in cryo-{EM}},
  author  = {Baldwin, Philip R. and Lyumkis, Dmitry},
  journal = {Progress in Biophysics and Molecular Biology},
  volume  = {150},
  pages   = {160--183},
  year    = {2020},
  doi     = {10.1016/j.pbiomolbio.2019.09.002}
}

@article{levy2025cryodrgnai,
  title   = {Revealing biomolecular structure and motion with neural ab initio cryo-{EM} reconstruction},
  author  = {Levy, Axel and Grzadkowski, Michal and Poitevin, Frederic and Vallese, Francesca and Clarke, Oliver B. and Wetzstein, Gordon and Zhong, Ellen D.},
  journal = {Nature Methods},
  year    = {2025},
  doi     = {10.1038/s41592-025-02720-4}
}

@inproceedings{jeon2024cryobench,
  title     = {{CryoBench}: diverse and challenging datasets for the heterogeneity problem in cryo-{EM}},
  author    = {Jeon, Minkyu and Raghu, Rishwanth and Astore, Miro A. and Woollard, Geoffrey and Feathers, Ryan and Kaz, Alkin and Hanson, Sonya M. and Cossio, Pilar and Zhong, Ellen D.},
  booktitle = {Advances in Neural Information Processing Systems (NeurIPS) Datasets and Benchmarks},
  year      = {2024}
}

@article{proteinrediff,
  author  = {Nguyen, Viet Thanh Duy and Nguyen, Nhan D. and Hy, Truong Son},
  title   = {{ProteinReDiff}: complex-based ligand-binding proteins redesign by equivariant diffusion-based generative models},
  journal = {Structural Dynamics},
  volume  = {11},
  number  = {6},
  pages   = {064102},
  year    = {2024},
  issn    = {2329-7778},
  doi     = {10.1063/4.0000271}
}

@article{liu2023kinase,
  author  = {Liu, Changchang and Kutchukian, Peter and Nguyen, Nhan D. and AlQuraishi, Mohammed and Sorger, Peter K.},
  title   = {A hybrid structure-based machine learning approach for predicting kinase inhibition by small molecules},
  journal = {Journal of Chemical Information and Modeling},
  volume  = {63},
  number  = {17},
  pages   = {5457--5472},
  year    = {2023},
  doi     = {10.1021/acs.jcim.3c00347}
}
%
%

\clearpage
\onecolumngrid

\begin{center}
  {\large\bfseries Supplementary Material}\\[4pt]
  {\normalsize for ``\maintitle''}\\[8pt]
  {\normalsize Nhan D. Nguyen and Bao Pham}
\end{center}
\vspace{1em}
\twocolumngrid

\appendix

\renewcommand{\thesection}{\Alph{section}}
\renewcommand{\thefigure}{S\arabic{figure}}
\renewcommand{\theequation}{\thesection\arabic{equation}}
\renewcommand{\thetable}{S\arabic{table}}
\renewcommand{\theHsection}{supp.\Alph{section}}
\renewcommand{\theHfigure}{supp.figure.\arabic{figure}}
\renewcommand{\theHequation}{supp.\Alph{section}.\arabic{equation}}
\renewcommand{\theHtable}{supp.table.\arabic{table}}

\setcounter{subsection}{0}
\setcounter{section}{0}
\setcounter{equation}{0}
\setcounter{table}{0}
\setcounter{figure}{0}

\onecolumngrid
\refstepcounter{table}\label{tab:notation}
\begingroup\footnotesize
\noindent\textbf{TABLE~\thetable.} Notation. Each symbol carries one meaning throughout; $L$ is reserved for the box side, $C$ for a channel count, and $\CTF$ for the contrast transfer function, following the convention of Refs.~\cite{singer2020computational,bendory2020cryoem}.

\vspace{2pt}
\noindent
\footnotesize
\begin{tabular}{lll}
\toprule
Symbol & Space & Meaning \\
\midrule
\multicolumn{3}{@{}l}{\emph{Forward model} (Sec.~\ref{sec:methods:forward})}\\
$L$, $C$ & $\mathbb{N}$ & box side in voxels; encoder input channels ($C=2$) \\
$\Vol$, $\hat{\Vol}$ & $\mathbb{R}^{L\times L\times L}$, $\mathbb{C}^{L\times L\times L}$ & volume and its Fourier transform \\
$y$, $\eta$ & $\mathbb{R}^{L\times L}$ & particle image; its additive noise \\
$\Rot$ & $\SO$ & particle orientation \\
$\xi$, $\CTF_\xi$ & $\mathbb{R}^{3}$, $\mathbb{R}^{L\times L}$ & per-particle defocus pair and astigmatism angle; the transfer function they define \\
$\Slice_{\Rot}$ & $\mathbb{C}^{L\times L\times L}\!\to\!\mathbb{C}^{L\times L}$ & central-slice operator \\
$\mathbf{q}$, $\mathbf{p}$ & $\mathbb{R}^{3}$ & Fourier coordinate in slice plane $\Pi_{\Rot}$; $\mathbf{p}=\Rot^{-1}\mathbf{q}$ \\
$k$, $\sigma^{2}(k)$ & $\mathbb{R}_{\ge0}$ & shell radius $\lVert\mathbf{q}\rVert$; noise power in that shell \\
\midrule
\multicolumn{3}{@{}l}{\emph{Information geometry} (Sec.~\ref{sec:methods:info})}\\
$\hat{e}$, $\hat{e}_j$ & $\mathbb{S}^{2}$ & rotation axis; the three body axes, $j=1,2,3$ \\
$\delta$, $\boldsymbol{\delta}$ & $\mathbb{R}$, $\mathbb{R}^{3}$ & rotation perturbation, as angle and as body-axis vector \\
$\mathcal{I}$, $\mathcal{I}_k$ & $\mathbb{R}_{>0}$ & Fisher information of a rotation component; its shell-$k$ part \\
$P_k$, $n_k$ & $\mathbb{R}_{>0}$, $\mathbb{N}$ & signal power in shell $k$; Fourier samples it holds \\
$R_g$, $\kappa$ & $\mathbb{R}_{>0}$ & radius of gyration about $\hat{e}$; geometric factor in Eq.~\eqref{eq:crb} \\
$\Lambda_k$, $c_k$ & $\mathbb{R}$ & shell-resolved score curvature; its correlation at the true pose \\
$\psi$, $\rho^{2}$ & $[0,\pi]$, $\mathbb{R}_{\ge0}$ & angle between $\hat{e}$ and $\mathbf{p}$; matched-filter statistic \\
\midrule
\multicolumn{3}{@{}l}{\emph{Estimator} (Secs.~\ref{sec:methods:model}, \ref{sec:methods:refine})}\\
$\mathcal{G}$, $\Delta$ & $\subset\SO$ & orientation grid, $\lvert\mathcal{G}\rvert=36{,}864$; its mean neighbor spacing \\
$M$, $B$, $N_{\mathrm{step}}$ & $\mathbb{N}$ & training structures ($3{,}330$); particles per training step ($96$); training steps ($2\times10^{6}$) \\
$\mathbf{t}_i$ & $\mathbb{R}^{L\times L}$ & template rendered at $\Rot_i\in\mathcal{G}$ \\
$f_\theta$, $f_\phi$ & $\to\mathbb{S}^{511}$ & particle and template encoders (unit-norm, $512$-dim) \\
$u_b$, $v_i$ & $\mathbb{S}^{511}$ & particle and template embeddings \\
$s(\Rot)$, $\tau$ & $\mathbb{R}$, $\mathbb{R}_{>0}$ & pose score, Eq.~\eqref{eq:logits}; learned temperature \\
$\zeta$ & $(0,1]$ & orientation information retained by phase flipping, Eq.~\eqref{eq:flipeff} \\
$\sigma_q$, $\epsilon$ & degrees & geodesic soft-target width; Newton probe angle \\
$g_j$, $\Lambda_j$ & $\mathbb{R}$ & first and second derivatives of $s$ along $\hat{e}_j$ \\
$\mathbf{g}$, $\mathbf{\Lambda}$ & $\mathbb{R}^{3}$, $\mathbb{R}^{3\times3}$ & gradient and Hessian of $s$ in the chart at $\Rot$ \\
$m$ & $\mathbb{N}$ & refinement iteration \\
\midrule
\multicolumn{3}{@{}l}{\emph{Denoising channel} (Sec.~\ref{sec:methods:diffusion})}\\
$t$, $T$ & $\{0,\dots,T\!-\!1\}$ & noise-level index; number of levels \\
$\hat{y}_0$, $y_t$, $\varepsilon_t$ & $\mathbb{C}^{L\times L}$ & clean, noised image; its noise ($\varepsilon_{T-1}\!=\!\eta$) \\
$\mathbf{A}_t$, $w_t$, $\bar{\alpha}_t$ & $\mathbb{R}^{L\times L}$, $[0,1]$ & transfer operator $\mathbf{1}\!\to\!\CTF$; its weight; noise schedule \\
$\varepsilon_\vartheta$, $\vartheta$ & $\to\mathbb{C}^{L\times L}$ & denoising U-Net and its parameters (distinct from $\theta$, $\phi$) \\
$\lambda$, $\gamma^{\star}$ & $\mathbb{R}_{>0}$ & Wiener regularizer $\sigma_n^{2}/\sigma_s^{2}$; the gain it induces \\
$\mathcal{L}_{\mathrm{CE}}$, $\mathcal{L}_{\varepsilon}$ & $\mathbb{R}$ & pose cross-entropy; $\varepsilon$-prediction loss, Eq.~\eqref{eq:epsloss} \\
\bottomrule
\end{tabular}
\endgroup
\vspace{1em}
\twocolumngrid


\newpage
\newpage


\begin{figure}[t]
\centering
\includegraphics[width=\linewidth,height=0.30\textheight,keepaspectratio]{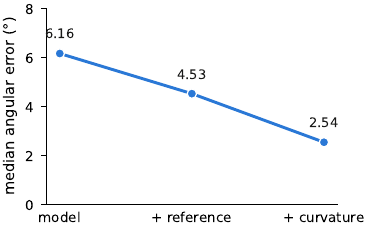}
\caption{\textbf{Where the accuracy is gained.} Median angular error on
EMPIAR-10076 through the refinement ladder: the grid maximum of the learned
posterior, re-scoring against a reference rebuilt from those poses, and the
curvature step.}
\label{fig:s1}
\end{figure}

\begin{figure}[t]
\centering
\includegraphics[width=\linewidth,height=0.30\textheight,keepaspectratio]{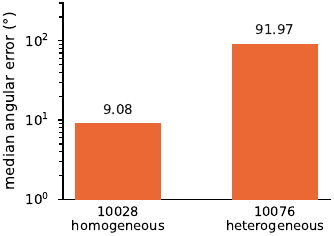}
\caption{\textbf{Set-size control for the ab-initio comparison.} CryoFastAR
evaluated on the homogeneous EMPIAR-10028 at the same particle count used for
EMPIAR-10076, isolating specimen heterogeneity from set size.}
\label{fig:s2}
\end{figure}

\begin{figure}[t]
\centering
\includegraphics[width=\linewidth,height=0.30\textheight,keepaspectratio]{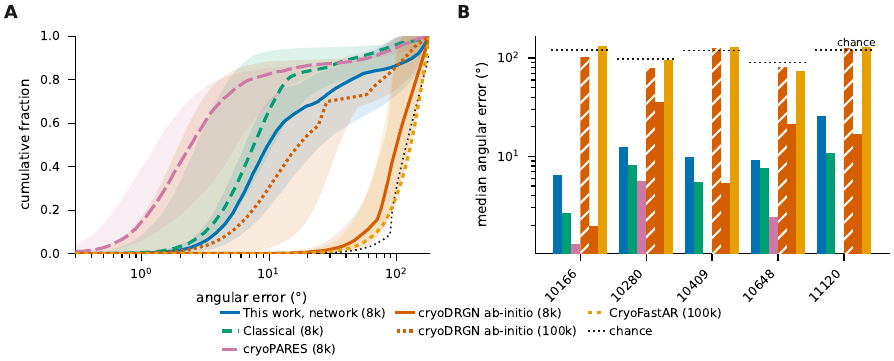}
\caption{\textbf{Reference-free and per-specimen baselines on matched particles.}
(A)~Cumulative angular error, median over the five CESPED targets with the full
range shaded. (B)~Median per target. All arms score the same $8{,}000$ particles
per target from one seeded selection; cryoPARES is matched to them by particle
identity (\texttt{rlnImageName}). Chance is drawn per target, since a point group
of order $\lvert G\rvert$ lets a random pose fall nearer an equivalent copy:
$\SI{120}{\degree}$ on the three C$_1$ targets, $\SI{97}{\degree}$ on the C$_2$
target, $\SI{90}{\degree}$ on the D$_2$ target.
cryoPARES was trained for four of the five targets and reaches $1.29$, $5.59$ and
$\SI{2.44}{\degree}$ on the matched particles. EMPIAR-11120 carries no cryoPARES
bar because no model exists for it; EMPIAR-10409 carries none in this panel
alone, where the name-matched subset was not computed, and its model reaches
$\SI{5.42}{\degree}$ on the unmatched $8{,}000$
(Tables~\ref{tab:cparestime}, \ref{tab:maptargets}, \ref{tab:transfer}).
The two reference-free methods solve a harder problem --- no reference and no
translations --- and ran at their default settings, each additionally given a
solved global rotation and a choice of handedness beyond the point-group
quotient every arm receives. Those allowances can only help them, and a
random-pose control through the identical alignment stays at
$\SI{119.6}{\degree}$, so the alignment manufactures no accuracy.
Particle count decides cryoDRGN. At $8{,}000$ particles it sits at chance on
every target; at $100{,}000$ it reaches $1.96$, $5.32$ and $\SI{17.09}{\degree}$
on the three C$_1$ targets, ahead of our network on all three, breaking from
chance only near epoch $14$ of $30$. It reaches $35.49$ (C$_2$) and
$\SI{21.53}{\degree}$ (D$_2$) on the symmetric targets, well above chance but far
behind the other arms, as \texttt{abinit\_homo} exposes no symmetry option. None
of these transfers: each is a separate optimization costing $2.6$--$5.3$
GPU-hours. CryoFastAR does not improve with count over the same range.}
\label{fig:s3}
\end{figure}

\section{Benchmark scoring details}\label{app:benchdetail}

\emph{Matched particles} (Fig.~\ref{fig:bench}A). Each estimator is scored on
data it did not train on. Our encoder and the classical filter never see any of
these proteins, so all $8{,}000$ particles per target are held out for both;
cryoPARES, being trained per specimen, is scored on the subset lying outside its
own training split. Restricting our two arms to that same subset moves their
medians by at most $\SI{0.09}{\degree}$, so the curves stay comparable.

\emph{Error bars} (Fig.~\ref{fig:bench}B). Bootstrap over particles, $200$
resamples.

\emph{Shell correlation} (Fig.~\ref{fig:bench}C). Every arm is reconstructed
from the same number of particles per half-set. The band is one standard
deviation across targets. The method curves are measured against the deposited
map, whereas \emph{half vs half} is our own two half-sets against each other.

\emph{Latent spread} (Fig.~\ref{fig:bench2}A). Each grey bar spans the strongest
to the weakest canonical direction at that dimension and the dots are the
individual directions, ranked. The spread is structure, not uncertainty ---
RECOVAR is deterministic given its inputs --- and it shows that the decline in
the mean is the trailing directions decorrelating rather than the leading one
degrading. The dashed line is the shuffled-pose floor.

\emph{View-sphere coverage} (Fig.~\ref{fig:bench2}B). The $20{,}000$
EMPIAR-10076 particles populate $637$ of the $648$ cells.

\emph{Ordering by mean error} (Fig.~\ref{fig:bench2}C). cryoPARES carries the
largest mean error of any arm shown while still producing the best map, because
its errors are tail-like rather than diffuse.

\begin{figure}[t]
\centering
\includegraphics[width=\linewidth,height=0.30\textheight,keepaspectratio]{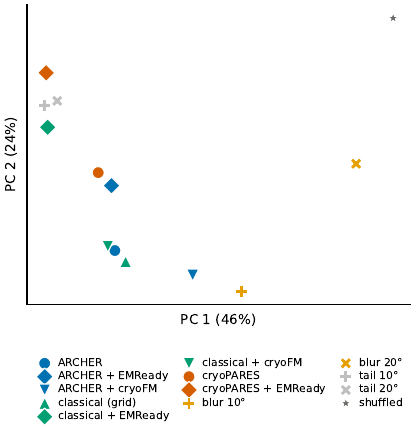}
\caption{\textbf{Reconstructions of EMPIAR-10409, by method.} Each arm's map of
that one specimen, embedded in the same principal-component space as the
projections of Fig.~\ref{fig:maps}B; one point is one map. Restored maps group
by restorer rather than by the estimator that produced the input, which is the
conclusion the displacement cosines of Fig.~\ref{fig:maps}C reach numerically.}
\label{fig:smethod}
\end{figure}

\section{Architecture and objective}\label{app:arch}

\subsection{Encoders}

Both encoders in Eq.~\eqref{eq:logits} are residual convolutional
networks\cite{he2016deep} with group normalization\cite{wu2018group} and SiLU
activations, mapping $(B, C, L, L) \to (B, 512)$ with an $\ell_2$-normalized
output. They share no weights, and share a topology apart from the stem's
input channels. The forward pass is, in order:

\begin{enumerate}\itemsep2pt
\item \textbf{Channel assembly.} The image branch of the reported model takes
  $C = 2$ channels, $[\,y,\ \mathcal{F}^{-1}[\mathrm{sign}(\CTF)\mathcal{F}y]\,]$;
  the template branch takes $C = 1$. The denoiser of
  Sec.~\ref{sec:methods:diffusion} supplies a third channel. Each channel is divided by its own standard deviation,
  so per-protein contrast does not enter the similarity. Wiener-corrected and
  whitened channels are implemented but are not part of the reported
  configuration.
\item \textbf{Stem.} One $5\times5$ convolution at stride 2, then SiLU. The
  stride is not cosmetic: one training step encodes hundreds of templates in one
  backward pass, and a stride-1 stem would hold every one of them at full
  $64\times64$ resolution simultaneously, which at the widths used here
  reaches gigabyte scale for a single activation tensor. Halving the
  resolution cuts that fourfold and costs little at $L = 64$, where the
  particle already spans most of the box.
\item \textbf{Residual trunk.} Four stride-2 blocks. Each block is
  \texttt{conv3$\times$3 $\to$ GroupNorm $\to$ SiLU $\to$ conv3$\times$3 $\to$
  GroupNorm}, added to a shortcut that is the identity when the shape is
  preserved and a $1\times1$ convolution otherwise, followed by SiLU. Group
  count is $\min(8, C_{\mathrm{out}})$. Channel width doubles per stage from a
  base width and is capped at four times that base, so the deepest stages do
  not dominate the parameter count. For the reported model the base width is
  128, giving stage widths $[128, 256, 512, 512]$ and a $2\times2$ trunk
  output at $L = 64$; the five-stage capacity arm of Table~\ref{tab:threearms}
  reaches $1\times1$.
\item \textbf{Projection head.} Global average pooling to a vector, then
  \texttt{Linear $\to$ SiLU $\to$ Linear} to dimension $512$, then $\ell_2$
  normalization onto $\mathbb{S}^{511}$.
\end{enumerate}

The logit scale $\tau^{-1}$ is a single learned scalar, stored as
$\log \tau^{-1}$. Because both embeddings are unit norm,
$\langle u, v_k\rangle \in [-1, 1]$, so $\tau^{-1}$ is the only quantity
setting how peaked the posterior in Eq.~\eqref{eq:logits} can become. It is
initialized at $\tau = 0.07$ and converges to $\tau = 0.0212$
($\tau^{-1} = 47.1$), a factor of $3.3$ sharper than initialization --- so the
scale is genuinely fitted rather than inherited. A clamp at
$\tau^{-1} \le 100$ guards against an early saturating softmax stalling the
gradient; it is never reached.

The reported model has $20.8$\,M parameters and a $512$-dimensional embedding; the capacity
ablation of Table~\ref{tab:threearms} uses a $69$\,M variant differing only in base
width, depth and embedding dimension.

\paragraph{What the architecture deliberately does not contain.} There is no
volume decoder, no latent variable, no sampling step, no attention and no
recurrence. The denoising channel is conditioned on a noise level and evaluated
in a single step, so it borrows the training objective of diffusion models
without being a generative model; the pose network itself generates nothing: it maps two images to two vectors and the pose posterior is
a softmax over their inner products against a \emph{fixed} candidate set. The restriction is deliberate: the volume given poses is a
closed-form linear solve (Eq.~\eqref{eq:wiener}), so a generative decoder would
have to re-learn a projection--slice relationship that is already known
exactly.

\subsection{CTF handling}

The contrast transfer function enters \emph{only} as pixel content, through the
phase-flipped second channel
$\mathcal{F}^{-1}[\mathrm{sign}(\CTF)\,\mathcal{F}y]$, and is never handed to
the network as a parameter vector: the network reads the transfer function in
the pixels it modulates, which is the same form it takes on a real micrograph.
That channel carries information rather than repeating the first --- measured on
real EMPIAR-10409 particles, the raw and phase-flipped channels correlate at
only $0.047$.

A Wiener-corrected channel is implemented alongside it, since phase flipping
fixes signs but not amplitudes, and it matters most where the CTF has no in-band
zero crossing: the $L = 64$ cache Fourier-crops every EMDB map to a fixed box,
so large particles land at coarse voxel sizes where the transfer function barely
oscillates. It is not part of the reported configuration.

\paragraph{The denoised channel.} The third-channel arm quoted in
Sec.~\ref{sec:methods:diffusion} shares every setting with the reported model
and differs only in the added channel, so the comparison of
Table~\ref{tab:denoisechan} is step-matched throughout. The gain is confined to
the fine rows: the lead on $<15^\circ$ is at chance, so the coarse assignment is
untouched and the channel acts on precision within the correct basin.

\begin{table}[ht]
\centering
\footnotesize
\caption{The denoised channel, step-matched. Both arms share every setting and
differ only in the added third channel. Means over the last $150$ step-matched
held-out evaluations, spanning steps $600{,}000$--$1{,}345{,}000$; the lead
column counts evaluations at which the $3$-channel arm is ahead. Consecutive
evaluations of one run are correlated, so those counts describe the trajectory
and are not independence tests. The accuracy figures reported in the main text
are the $2$-channel model.}
\label{tab:denoisechan}
\fitcol{%
\begin{tabular}{@{}lrrr@{}}
\toprule
 & 3-channel & 2-channel & leads \\
\midrule
Median error      & \textbf{5.07}$^\circ$ & 5.22$^\circ$ & 122/150 \\
Within $5^\circ$  & \textbf{0.493} & 0.474 & 121/150 \\
Within $15^\circ$ & 0.755 & 0.753 & 85/150 \\
\bottomrule
\end{tabular}}
\end{table}

\subsection{Objective and optimization}

Training minimizes the soft cross-entropy of Eq.~\eqref{eq:logits}. The
soft-target width is tied to the grid: $\sigma_q = 6^\circ$ for the reported
order-3 model on its $7.40^\circ$ spacing, and $\sigma_q = 12^\circ$ on the
$14.78^\circ$ order-2 grid, i.e.\ $\sigma_q \approx 0.81\,\Delta$
in both cases. Where a hard target is used instead, label smoothing of $0.05$
is applied, matching the reference $\SO$-classifier
implementation.\cite{sanchezgarcia2025cryopares,szegedy2016rethinking}

Optimization uses AdamW with decoupled weight
decay\cite{loshchilov2019decoupled} of $1\times10^{-5}$, held constant across
every checkpoint in this work, on a cosine schedule with warm
restarts.\cite{loshchilov2017sgdr} The reported model trains at a learning
rate of $1\times10^{-4}$ with $T_{\max} = 200{,}000$. The short 150k-step
order-2 runs of Table~\ref{tab:threearms} use $3\times10^{-4}$; that value was
inherited by the $69$\,M capacity arm from its $20.8$\,M control rather than
retuned for it, which is the stated scope limit on that comparison.

\paragraph{One schedule property worth stating.} Setting \texttt{T\_max} equal
to the step budget anneals the learning rate to zero exactly while the model is
still improving, so a fixed-budget figure is a lower bound set by that budget
rather than a converged value. Reported figures are correspondingly plateau
medians over a stated window, and any continuation restarts the schedule.

\section{Full ablation and sweep tables}\label{app:tables}

Throughout this section \textbf{bold} marks the best value in a column wherever
the column holds directly comparable entries; tables that locate an operating
point rather than compare methods carry no emphasis, and the setting adopted is
named in the caption.

\subsection{Which configuration produced which number}\label{app:whichnumber}

Table~\ref{tab:map} states, for every accuracy figure reported anywhere in this
work, the grid order, the evaluation noise range, the particle count, the number
of proteins and whether point-group symmetry was quotiented. One difference
between configurations dominates the rest. The main-text held-out figures are
measured on the order-3 grid ($36{,}864$ cells, $\Delta = 7.40^\circ$, soft
target $\sigma_q = 6^\circ$), while the ablation and curriculum tables below
(Tables~\ref{tab:stages}--\ref{tab:matched} and \ref{tab:symrescore}) are sweeps
over training choices and were run on the cheaper order-2 grid ($4{,}608$ cells,
$\Delta = 14.78^\circ$, $\sigma_q = 12^\circ$). Halving the grid spacing roughly
halves the median error, so the order-2 medians of $10$--$\SI{12}{\degree}$ and
the order-3 median of $\SI{5.0}{\degree}$ are one model family measured at two
resolutions. Nothing in this work compares an order-2 number against an order-3
one.

\begin{table}[ht]
\centering
\footnotesize
\caption{Every reported accuracy figure and the configuration that produced it.
\emph{Grid} is the HEALPix order and the resulting cell count; \emph{SNR} is the
evaluation noise range, pinned explicitly whenever checkpoints are compared;
\emph{sym.} is whether the geodesic is minimized over the point group. The
order-2 rows are training-choice sweeps, not the reported model.}
\label{tab:map}
\fitcol{%
\begin{tabular}{@{}llllrrl@{}}
\toprule
Figure & Where & Grid & SNR & part. & prot. & sym. \\
\midrule
\multicolumn{7}{@{}l}{\textit{Held-out synthetic proteins}}\\
median $5.0^\circ$; $<15^\circ$ $0.77$; top-1 $0.41$
  & \S\ref{sec:res:amortization} & 3 ($36{,}864$) & $0.02$--$0.30$ & 768 & 100 & no \\
$<15^\circ$ $0.6807$; median $10.07^\circ$
  & Table~\ref{tab:stages}, \ref{tab:threearms} & 2 ($4{,}608$) & $0.02$--$0.30$ & 960 & 100 & no \\
hard-noise $<15^\circ$ $0.4047$
  & Table~\ref{tab:stages} & 2 ($4{,}608$) & hard & 960 & 100 & no \\
$<15^\circ$ $0.5463$; sym-aware $0.6180$
  & Table~\ref{tab:symrescore} & 2 ($4{,}608$) & $0.02$--$0.30$ & 192 & 100 & both \\
classical $<15^\circ$ $0.6412$--$0.9161$
  & Table~\ref{tab:sweep} & 2 ($4{,}608$) & swept & 192 & 30 & no \\
\midrule
\multicolumn{7}{@{}l}{\textit{Experimental particles}}\\
median $\SI{2.5}{\degree}$ (EMPIAR-10076)
  & \S\ref{sec:res:amortization} & 3 ($36{,}864$) & --- & $20{,}000$ & 1 & C$_1$ \\
CESPED transfer
  & Fig.~\ref{fig:bench}A & 3 ($36{,}864$) & --- & $8{,}000$ & 5 & yes \\
$\rho^{2}$, $<15^\circ$ $0.8217$
  & Table~\ref{tab:rho2} & 2 ($4{,}608$) & --- & $2{,}000$ & 1 & no \\
\bottomrule
\end{tabular}}
\end{table}

\subsection{The anneal-endpoint dose--response}

Table~\ref{tab:stages} gives the per-stage detail behind
Sec.~\ref{sec:methods:forward}. The hard-noise metric improves monotonically across all five stages, and the
legacy metric across the first four before flattening ($0.6141$ to $0.6130$ at
the last), so the trend is a dose--response rather than a single point.

\begin{table}[ht]
\centering
\footnotesize
\caption{Re-annealing to the calibrated endpoint. The legacy metric is held at
the pinned $(0.02, 0.30)$ range in every row, so the rows are directly
comparable. Stages are numbered along the geometric anneal (ratio
$\approx 0.665$ per stage); the warm start enters at stage 6. The stage-10 row
is a plateau median over the 12 evaluations of that stage (steps
350k--377.5k). The cycle-2 row continues the stage-10 arm through a
learning-rate restart rather than standing beside it, so it is set below the
rule; \textbf{that row is the reported model}.}
\label{tab:stages}
\fitcol{%
\begin{tabular}{@{}lrrrr@{}}
\toprule
Stage & mean SNR & legacy $<15^\circ$ & median & hard-noise $<15^\circ$ \\
\midrule
Warm start (167.5k) & 1.60 & 0.4688 & 16.8$^\circ$ & --- \\
6  & 0.206 & 0.5042 & 14.7$^\circ$ & 0.1719 \\
7  & 0.137 & 0.5573 & 12.9$^\circ$ & 0.2255 \\
8  & 0.091 & 0.5979 & 11.7$^\circ$ & 0.2766 \\
9  & 0.060 & 0.6141 & 11.4$^\circ$ & 0.3135 \\
10 (target, plateau) & 0.040 & 0.6130 & 11.4$^\circ$ & 0.3260 \\
\midrule
\multicolumn{5}{@{}l}{\textit{After the cycle-2 learning-rate restart}} \\
Cycle 2 plateau & 0.040 & \textbf{0.6807} & \textbf{10.07$^\circ$} & \textbf{0.4047} \\
\bottomrule
\end{tabular}}
\end{table}

\subsection{All three arms at their own plateaus}

Table~\ref{tab:threearms} reports the three endpoint arms, each at its own
plateau criterion. Table~\ref{tab:matched} is the controlled version of the same
question: there both arms resume the \emph{same} checkpoint under the same
schedule and differ only in the SNR bounds.

\begin{table}[ht]
\centering
\footnotesize
\caption{Endpoint arms at their own plateau criteria. Arm~B is a from-scratch
replication at the calibrated endpoint; Arm~C is warm-started at the same
endpoint; Arm~A is warm-started at the legacy endpoint. Bold marks the best of
the three. The reported model continues Arm~C through a learning-rate restart
and exceeds all of them (Table~\ref{tab:stages}, last row).}
\label{tab:threearms}
\fitcol{%
\begin{tabular}{@{}lrrr@{}}
\toprule
 & B (scratch) & C (warm) & A (warm) \\
 & calibrated & calibrated & legacy \\
\midrule
Legacy $<15^\circ$   & \textbf{0.6542} & 0.6130 & 0.5922 \\
Legacy $<30^\circ$   & \textbf{0.7219} & 0.6849 & 0.6781 \\
Median error         & \textbf{10.5$^\circ$} & 11.4$^\circ$ & 12.0$^\circ$ \\
Hard-noise $<15^\circ$ & \textbf{0.3760} & 0.3260 & 0.2651 \\
Total steps          & 462.5k & 377.5k & 397.5k \\
Real-particle $<15^\circ$ & \textbf{0.4358} & 0.3938 & 0.2727 \\
\bottomrule
\end{tabular}}
\end{table}

\begin{table}[ht]
\centering
\footnotesize
\caption{The controlled endpoint comparison. Both arms resume the \emph{same}
checkpoint with the same stage count, caps, plateau window, patience,
learning-rate schedule and plateau metric; the only difference is the SNR
bounds. Medians over the last 10 matched-step evaluations, from 32 matched
steps spanning 170k--247.5k. The calibrated arm leads at 32 of 32 matched
steps on both metrics. Consecutive evaluations of one run are correlated, so
that count is reported as a description of the trajectory and not as an
independence test.}
\label{tab:matched}
\fitcol{%
\begin{tabular}{@{}lrr@{}}
\toprule
Endpoint & legacy $<15^\circ$ & hard-noise $<15^\circ$ \\
\midrule
Control, legacy $(0.02, 0.30)$ & 0.5156 & 0.1677 \\
Calibrated $(0.005, 0.075)$ & \textbf{0.5573} & \textbf{0.2255} \\
Difference & $+$0.0417 & $+$0.0578 \\
\bottomrule
\end{tabular}}
\end{table}

\subsection{Detectability of real versus synthetic particles}

Table~\ref{tab:rho2} is the measurement establishing that experimental
particles carry ample pose information at this box size. A single real
particle sits $1.7\times$ above the $2\ln M = 16.9$ detection threshold and
the median particle's true pose ranks first of $4{,}608$.

\begin{table}[ht]
\centering
\footnotesize
\caption{Whitened matched-filter detectability, CESPED EMPIAR-10409 at
$L = 64$ ($3.128\,\Ang$/px, Nyquist $6.26\,\Ang$) against a reconstructed
reference. \emph{rank} is the rank of the true pose among $M = 4{,}608$
candidates for the median particle. The real-particle row is the anchor the
noise calibration targets; the synthetic rows bracket it. Real particles are
\texttt{rlnRandomSubset == 2} scored against the half-1 map, so none
contributed to the reference; the partly-overlapping variant gives $0.8380$ at
$\rho^2 = 29.44$, a 2\% difference.}
\label{tab:rho2}
\fitcol{%
\begin{tabular}{@{}lrrrrr@{}}
\toprule
Arm & $\rho^2$ & true-pose $z$ & rank & $<15^\circ$ & median \\
\midrule
Real (independent half) & 28.84 & $+$4.233 & 0 & 0.8217 & 8.1$^\circ$ \\
Real (shuffled poses) & 3.22 & $+$0.068 & 2159 & 0.0063 & 129.0$^\circ$ \\
Synthetic, SNR 0.03 & 40.85 & $+$4.955 & 0 & 0.9527 & 7.8$^\circ$ \\
Synthetic, SNR 0.01 & 13.76 & $+$3.341 & 2 & 0.5303 & 13.2$^\circ$ \\
Synthetic, SNR 0.003 & 4.14 & $+$1.962 & 115 & 0.0850 & 120.9$^\circ$ \\
\bottomrule
\end{tabular}}
\end{table}

\subsection{Rejecting junk particles with the pose confidence}\label{app:junk}

The posterior the pipeline already computes carries a usable quality signal at
no extra cost. We score the normalized negative entropy
$1 + \sum_i p_i \log p_i / \log \lvert\mathcal{G}\rvert$, the same confidence
the refinement stage uses, and ask how well it separates real particles from
junk. CESPED ships curated particles, so junk is constructed, and the
construction is the substance of the test: \emph{phase scramble} randomizes
Fourier phases at the measured amplitude spectrum, so every shell carries
identical power and only structure distinguishes it; \emph{pure noise} redraws
amplitudes from the mean shell profile, giving an image with no particle in it;
\emph{off-center} translates a real particle $25$--$45\%$ of the box, the bad
pick that actually populates real datasets. Separation is reported as
AUROC, the probability that a real particle outscores a junk one. The reference
is built from half~1 and the scored particles are drawn from half~0, so the
confidence cannot be reading back its own map.

\begin{table}[ht]
\centering
\footnotesize
\caption{Junk rejection on EMPIAR-10409, $2{,}000$ particles at $L=64$ against
a half-1 reference, for the reported model. AUROC of the pose confidence, real
versus each junk class; $0.5$ is no separation. Median confidence is $0.364$ on
real particles against $0.16$--$0.18$ on junk. The phase-scrambled row is the
strongest control, since it matches the real amplitude spectrum shell for
shell.}
\label{tab:junk}
\fitcol{%
\begin{tabular}{@{}lr@{}}
\toprule
Junk class & AUROC \\
\midrule
Phase scramble (matched spectrum) & 0.9626 \\
Pure noise & 0.9690 \\
Off-center pick & 0.9630 \\
\bottomrule
\end{tabular}}
\end{table}

The reported model separates all three classes at AUROC $0.963$--$0.969$. The phase-scrambled control is the informative one: an
image with the real power spectrum and destroyed structure is rejected as
readily as pure noise, so the score is responding to particle structure rather
than to a difference in image statistics that any variance threshold would
catch.

\subsection{The behavioral calibration sweep and the symmetry split}

Table~\ref{tab:sweep} gives the sweep behind Sec.~\ref{sec:methods:forward}.
Averaged over all held-out proteins the classical scorer settles near the
ceiling that ambiguity alone implies (Table~\ref{tab:ambiguity}), well short of
the $0.8217$ it achieves on real particles. Restricted to non-symmetric
proteins it reaches $0.8222$ at SNR $0.04$, matching the real-particle anchor
to within $0.0005$. The averaged wall is therefore a property of pooling
proteins whose poses are not separable, not of the scorer.

\begin{table}[ht]
\centering
\footnotesize
\caption{Classical $<15^\circ$ against synthetic SNR, split by whether the
protein admits indistinguishable poses. The all-proteins column saturates at
the ceiling implied by ambiguity alone; the non-symmetric column crosses the
real-data anchor of $0.8217$ at SNR $0.04$. Bold marks that row, the calibrated
endpoint adopted throughout; it is the operating point, not the largest entry.}
\label{tab:sweep}
\fitcol{%
\begin{tabular}{@{}lrrr@{}}
\toprule
Synthetic SNR & all proteins & non-symmetric & symmetric \\
\midrule
0.30 & 0.7094 & 0.9161 & 0.2490 \\
0.08 & 0.6905 & 0.8960 & 0.2317 \\
\textbf{0.04} & 0.6412 & 0.8222 & 0.2141 \\
0.02 & 0.5115 & 0.6361 & 0.1700 \\
\bottomrule
\end{tabular}}
\end{table}

\subsection{Symmetry: the ambiguity audit and its point-group verification}

Table~\ref{tab:ambiguity} gives the ambiguity statistics on the 30-protein
sweep set, where the predicted ceiling and the observed classical saturation
agree to 0.2\%. Table~\ref{tab:pointgroup} gives the quantization check on the
full 100-protein held-out set. Every one of the 26 symmetric proteins lands on
an integer point-group order with none scattered in between, which a
correlation artefact could not produce; the dominance of C2 and D2/C4 is what
an EMDB-derived corpus of oligomeric complexes should look like.

\begin{table}[ht]
\centering
\footnotesize
\caption{Pose ambiguity measured from the volumes alone --- no particles, no
noise, no scorer.}
\label{tab:ambiguity}
\fitcol{%
\begin{tabular}{@{}lr@{}}
\toprule
Statistic (30-protein sweep set) & Value \\
\midrule
Mean fraction of poses that are ambiguous & 0.3000 \\
Proteins with $>50\%$ ambiguous poses & 9 of 30 \\
corr(ambiguous fraction, classical accuracy) & $-$0.8431 \\
corr(voxel size, classical accuracy) & $+$0.2922 \\
Ceiling implied by ambiguity alone & 0.7000 \\
Classical accuracy actually observed & 0.6985 \\
\midrule
\multicolumn{2}{@{}l}{\textit{Extended to all 100 held-out proteins}} \\
Symmetric proteins & 26 \\
Proteins at exactly zero ambiguity & 71 \\
Proteins in between & 3 \\
Achievable ceiling & 0.7388 \\
corr(ambiguity, voxel size) & 0.0124 \\
\bottomrule
\end{tabular}}
\end{table}

\begin{table}[ht]
\centering
\footnotesize
\caption{Independent verification of the ambiguity measurement by point-group
quantization. A point group of order $|G|$ predicts exactly $|G|-1$
indistinguishable alternatives per pose.}
\label{tab:pointgroup}
\fitcol{%
\begin{tabular}{@{}rrrl@{}}
\toprule
Indistinguishable alternatives & Proteins & Implied $|G|$ & Point group \\
\midrule
1 & 10 & 2 & C2 \\
3 & 12 & 4 & C4 / D2 \\
6 & 1  & 7 & C7 \\
7 & 3  & 8 & C8 / D4 \\
\bottomrule
\end{tabular}}
\end{table}

\begin{table}[ht]
\centering
\footnotesize
\caption{Re-scoring both checkpoints with symmetry separated out. All 100
held-out proteins, 192 particles each, both checkpoints scored at the
\emph{same} pinned $(0.02, 0.30)$ range. Arrows run from the 167.5k checkpoint
to the plateaued calibrated checkpoint. The non-symmetric row is the control
confirming the symmetry-aware correction is not simply inflationary: it moves
almost nothing there.}
\label{tab:symrescore}
\fitcol{%
\begin{tabular}{@{}lrrr@{}}
\toprule
Subset & $n$ & $<15^\circ$ & symmetry-aware $<15^\circ$ \\
\midrule
All held-out proteins & 100 & 0.4223 $\to$ \textbf{0.5463} & 0.4771 $\to$ \textbf{0.6180} \\
Non-symmetric only & 74 & 0.5255 $\to$ \textbf{0.6779} & 0.5393 $\to$ \textbf{0.6892} \\
Symmetric only & 26 & 0.1284 $\to$ \textbf{0.1719} & 0.3001 $\to$ \textbf{0.4155} \\
\bottomrule
\end{tabular}}
\end{table}

\subsection{Per-target accuracy against the per-specimen estimator}\label{app:cpunseen}

Table~\ref{tab:cpunseen} is the per-target breakdown behind the comparison in
Sec.~\ref{sec:res:amortization}. cryoPARES is scored only where it did not
train, and all four arms are scored on exactly those particles, so no arm is
credited for particles another arm never saw. Pooling the table by taking the
median across the four targets gives medians of \SI{9.5}{\degree} for the
network, \SI{7.5}{\degree} for our pipeline, \SI{6.6}{\degree} classical and
\SI{4.2}{\degree} for cryoPARES, and fractions within \SI{15}{\degree} of
$0.689$, $0.714$, $0.809$ and $0.805$. The two statistics do not induce the same
ordering: cryoPARES leads on the median by a wide margin and ties the classical
filter on the tail.

\begin{table}[ht]
\centering
\footnotesize
\caption{Per-target accuracy on the particles cryoPARES did not train on.
All four arms are scored on the \emph{same} particles of each target --- the
intersection of our seeded $8{,}000$ with the complement of that target's
cryoPARES training split, $n$ below --- with the same symmetry-aware geodesic,
so the columns are directly comparable. Bold marks the best arm per target.
cryoPARES holds the best median on three of the four targets while the
classical filter holds the largest fraction within $\SI{15}{\degree}$ on two of
them: the two statistics rank the arms differently, which is why both are
reported. EMPIAR-10409 is the clearest case: cryoPARES leads on neither
statistic there, taking a median within $\SI{0.15}{\degree}$ of the classical
filter while placing $0.179$ fewer particles within $\SI{15}{\degree}$ --- a
tail close to that of our unrefined grid argmax.}
\label{tab:cpunseen}
\fitcol{%
\begin{tabular}{@{}lrrrr@{}}
\toprule
 & EMPIAR-10166 & EMPIAR-10280 & EMPIAR-10409 & EMPIAR-10648 \\
\midrule
$n$ scored & 1269 & 4562 & 4079 & 1190 \\
\midrule
\multicolumn{5}{@{}l}{\textit{Median angular error (degrees)}} \\
Network (grid argmax) & 6.28 & 12.48 & 9.91 & 9.04 \\
\methodname{} pipeline & 2.97 & 10.14 & 7.06 & 7.90 \\
Classical matched filter & 2.63 & 8.13 & \textbf{5.54} & 7.60 \\
cryoPARES & \textbf{1.46} & \textbf{5.79} & 5.69 & \textbf{2.61} \\
\midrule
\multicolumn{5}{@{}l}{\textit{Fraction within $15^\circ$}} \\
Network (grid argmax) & 0.7896 & 0.5701 & 0.6249 & 0.7538 \\
\methodname{} pipeline & 0.8046 & 0.6223 & 0.6506 & 0.7782 \\
Classical matched filter & \textbf{0.9409} & 0.7429 & \textbf{0.8120} & 0.8067 \\
cryoPARES & 0.8448 & \textbf{0.7646} & 0.6330 & \textbf{0.8462} \\
\bottomrule
\end{tabular}}
\end{table}

\begin{table}[ht]
\centering
\footnotesize
\caption{Training cost of the per-specimen comparison. cryoPARES
\cite{sanchezgarcia2025cryopares} is supervised on the target it is applied to,
so each entry is a separate optimization; times are single-GPU wall clock for
the completed run on each dataset at the scope used here ($200{,}000$ particles,
$40$ epochs). \methodname{} is trained once across $3{,}330$ structures and
applied to every target without retraining, so its cost does not recur per
specimen --- for a new target it is an inference pass.}
\label{tab:cparestime}
\fitcol{%
\begin{tabular}{@{}lr@{}}
\toprule
Target & GPU-hours \\
\midrule
EMPIAR-10166 & 13.2 \\
EMPIAR-10280 & 22.8 \\
EMPIAR-10409 & 23.3 \\
EMPIAR-10648 & 31.2 \\
\midrule
Total, four targets & 90.5 \\
\bottomrule
\end{tabular}}
\end{table}

\begin{table}[ht]
\centering
\footnotesize
\caption{Reconstruction quality per target against the deposited map, as the
unmasked $0.143$ crossing. Unmasked is reported because the masked curve does
not cross before Nyquist on EMPIAR-10409 and 10648, so a masked number there is
a floor rather than a measurement (Fig.~\ref{fig:physics}C). Our poses were
predicted for $60{,}000$ particles per half on EMPIAR-10166 and 10280 against
cryoPARES's $100{,}000$, so on those two we also rebuilt cryoPARES from the same
$60{,}000$ (last column). Matching the count recovers $0.49$ and
$\SI{0.37}{\angstrom}$ of the gap and cryoPARES still leads by $1.62$ and
$\SI{1.69}{\angstrom}$, so the deficit there is mostly pose quality and not set
size. The reverse control --- rebuilding cryoPARES at the \emph{larger} counts
we used on EMPIAR-10409 and 10648 --- was not run, hence the two dashes; it
would widen the gap on those targets rather than narrow it. Entries are rounded
to two decimals, which is coarser than the differences discussed in the main
text: on EMPIAR-10409 the three values are $3.634$ (ours), $3.625$ (classical)
and $\SI{3.477}{\angstrom}$ (cryoPARES), a spread of
$\SI{0.157}{\angstrom}$. The classical arm, reconstructed from the same particles as ours with no
learned component, lands within $\SI{0.3}{\angstrom}$ of our pipeline on every
target: the gap to cryoPARES is therefore a property of reference-conditioned
refinement at this working resolution and not of the network. On the two targets where we supply \emph{more} particles than cryoPARES
the gap is $\SI{0.16}{\angstrom}$ and $\SI{0.22}{\angstrom}$; the former is the
agreement quoted in the main text.}
\label{tab:maptargets}
\fitcol{%
\begin{tabular}{@{}lrrrrrr@{}}
\toprule
 & \multicolumn{2}{c}{particles per half} & \multicolumn{4}{c}{resolution (\Ang)} \\
\cmidrule(lr){2-3}\cmidrule(lr){4-7}
Target & ours & cryoPARES & ours & classical & cryoPARES & cP at our count \\
\midrule
EMPIAR-10166 & 60{,}000 & 100{,}000 & 5.97 & 5.91 & 3.86 & 4.35 \\
EMPIAR-10280 & 60{,}000 & 100{,}000 & 6.30 & 6.14 & 4.24 & 4.61 \\
EMPIAR-10409 & 203{,}000 & 100{,}000 & 3.63 & 3.62 & 3.48 & --- \\
EMPIAR-10648 & 117{,}478 & 100{,}000 & 4.27 & 4.56 & 4.05 & --- \\
\bottomrule
\end{tabular}}
\end{table}

\begin{table}[ht]
\centering
\footnotesize
\caption{cryoPARES transfer. Each row is a model trained on one target; each
column the particles it is applied to. Values are median angular error in
degrees on the same $8{,}000$ particles per target, symmetry-aware, using the
network stage alone --- the local-refinement stage projects the training
target's own reference volume and cannot be run across specimens. The diagonal
(bold) is the same-specimen setting every other cryoPARES number in this work
reports. Chance is $\SI{120}{\degree}$ for the C$_1$ targets 10166 and 10409,
$\SI{97}{\degree}$ for C$_2$ 10280 and $\SI{90}{\degree}$ for D$_2$ 10648, so
every off-diagonal entry is at chance.}
\label{tab:transfer}
\fitcol{%
\begin{tabular}{@{}lrrrr@{}}
\toprule
 & \multicolumn{4}{c}{particles from} \\
\cmidrule(l){2-5}
model trained on & 10166 & 10280 & 10409 & 10648 \\
\midrule
10166 & \textbf{3.9} & 104.0 & 124.6 & 83.6 \\
10280 & 133.2 & \textbf{7.8} & 132.1 & 86.0 \\
10409 & 134.9 & 107.3 & \textbf{8.8} & 80.2 \\
10648 & 129.5 & 97.2 & 134.7 & \textbf{5.4} \\
\midrule
chance & 120 & 97 & 120 & 90 \\
\bottomrule
\end{tabular}}
\end{table}

\section{Verification of the geometric conventions}\label{app:validation}

Table~\ref{tab:verify} lists the checks that fix the sign, centering and
transfer-function conventions used throughout. Each compares an implementation
against an expectation derived outside it --- an analytic identity, a reference
implementation, or an independent reconstructor --- so agreement is evidence
rather than self-consistency.

\begin{table}[ht]
\centering
\footnotesize
\caption{Geometric and CTF conventions checked against external references;
all five agree. Real particles
backprojected at their deposited poses reconstruct to the $L = 64$ Nyquist limit
with mean FSC $0.9325$ over shells 3--20 against a shuffled-pose floor of
$0.0947$; the external anchor is an independent codebase's reconstructor at
$4.17\,\Ang$ from the same poses and particles.}
\label{tab:verify}
\begingroup
\newcommand{\tblkey}[1]{\begin{minipage}[t]{0.20\textwidth}\raggedright #1\strut\end{minipage}}
\newcommand{\tblval}[1]{\begin{minipage}[t]{0.24\textwidth}\raggedright #1\strut\end{minipage}}
\fitcol{%
\begin{tabular}{@{}ll@{}}
\toprule
\tblkey{Check} & \tblval{Result} \\
\midrule
\tblkey{Checkerboard $\equiv$ \texttt{fftshift}} & \tblval{max abs diff $0.00\mathrm{e}{+}00$} \\
\tblkey{In-plane composition identity} & \tblval{centered $+0.9888$ vs uncentered $+0.0535$} \\
\tblkey{CTF sign vs reference implementation} & \tblval{phase-flip sign agreement $1.0000$} \\
\tblkey{Slicer/backprojector round trip} & \tblval{true $0.9857$ vs shuffled $0.1168$} \\
\tblkey{Real particles at deposited poses} & \tblval{$6.3\,\Ang$, FSC $0.9325$ vs shuffled $0.0947$} \\
\bottomrule
\end{tabular}}
\endgroup
\end{table}

\section{Reproducibility settings}\label{app:repro}

\begin{table}[ht]
\centering
\footnotesize
\caption{Settings for the reported model and its evaluation.}
\label{tab:repro}
\begingroup
\newcommand{\tblkey}[1]{\begin{minipage}[t]{0.20\textwidth}\raggedright #1\strut\end{minipage}}
\newcommand{\tblval}[1]{\begin{minipage}[t]{0.24\textwidth}\raggedright #1\strut\end{minipage}}
\fitcol{%
\begin{tabular}{@{}ll@{}}
\toprule
\tblkey{Setting} & \tblval{Value} \\
\midrule
\multicolumn{2}{@{}l}{\textbf{Corpus and splits}} \\
\tblkey{Training structures} & \tblval{3{,}330 EMDB entries} \\
\tblkey{Held-out structures} & \tblval{100, split by structure ID} \\
\tblkey{Split granularity} & \tblval{structure, never particle} \\
\tblkey{Voxel-size range} & \tblval{median 4.7\,\Ang/px; 90\% within 3.1--6.9\,\Ang/px} \\
\tblkey{Box size} & \tblval{$L = 64$} \\
\addlinespace
\multicolumn{2}{@{}l}{\textbf{Model}} \\
\tblkey{Parameters} & \tblval{20.8\,M (capacity arm: 69\,M)} \\
\tblkey{Embedding dimension} & \tblval{$512$} \\
\tblkey{Input channels} & \tblval{[raw, phase-flipped]; third: denoised} \\
\addlinespace
\multicolumn{2}{@{}l}{\textbf{Pose grid}} \\
\tblkey{HEALPix order} & \tblval{2 ($M = 4{,}608$); 3 ($M = 36{,}864$)} \\
\tblkey{Measured spacing $\Delta$} & \tblval{$14.78^\circ$ at $o=2$; $7.40^\circ$ at $o=3$} \\
\tblkey{Half-spacing $\Delta/2$} & \tblval{$7.39^\circ$ at $o=2$; $3.70^\circ$ at $o=3$} \\
\addlinespace
\multicolumn{2}{@{}l}{\textbf{Objective and optimization}} \\
\tblkey{Soft-target width} & \tblval{$\sigma = 6^\circ$ at $o=3$; $12^\circ$ at $o=2$ ($\approx 0.81\,\delta_{\mathrm{grid}}$)} \\
\tblkey{Label smoothing (hard-target variant)} & \tblval{0.05} \\
\tblkey{Optimizer} & \tblval{AdamW, weight decay $1\times10^{-5}$} \\
\tblkey{Learning rate, reported model} & \tblval{$1\times10^{-4}$, cosine with warm restarts, $T_{\max} = 200{,}000$} \\
\tblkey{Learning rate, 150k order-2 runs} & \tblval{$3\times10^{-4}$ (also used, unretuned, by the 69\,M arm)} \\
\tblkey{Logit scale $\tau^{-1}$} & \tblval{learned; init $\tau = 0.07$, converged $\tau = 0.0212$; clamp $\le 100$ not reached} \\
\tblkey{Particles per training step} & \tblval{96} \\
\tblkey{Reference run length} & \tblval{150k steps ($\sim$3--6\,h, one A100)} \\
\tblkey{Reported model budget} & \tblval{$2\times10^{6}$ steps} \\
\addlinespace
\multicolumn{2}{@{}l}{\textbf{Noise curriculum}} \\
\tblkey{Legacy anneal endpoint} & \tblval{$(0.02, 0.30)$, mean SNR 0.16} \\
\tblkey{Calibrated anneal endpoint} & \tblval{$(0.005, 0.075)$, mean SNR 0.04} \\
\tblkey{Hard-noise eval range} & \tblval{$(0.0023, 0.0348)$, mean SNR 0.019} \\
\tblkey{Noise model} & \tblval{measured 2D PSD, angular structure retained} \\
\addlinespace
\multicolumn{2}{@{}l}{\textbf{Real-particle evaluation}} \\
\tblkey{Primary dataset} & \tblval{CESPED EMPIAR-10409, $3.128\,\Ang$/px, Nyquist $6.26\,\Ang$} \\
\tblkey{Additional datasets} & \tblval{EMPIAR-10166, 11120, 10280, 10648} \\
\tblkey{Reference/scoring split} & \tblval{score subset~2 against a subset-1 reference} \\
\tblkey{$\rho^2$ band} & \tblval{7.1--100.1\,\Ang} \\
\tblkey{FSC criterion} & \tblval{0.143, against an independent-half reference} \\
\addlinespace
\multicolumn{2}{@{}l}{\textbf{Reporting protocol}} \\
\tblkey{Training figures} & \tblval{plateau median over last 40\% of evaluations} \\
\tblkey{Minimum window} & \tblval{$\sim$10 evaluations} \\
\tblkey{Cross-checkpoint comparisons} & \tblval{evaluation noise range pinned explicitly} \\
\tblkey{Minimum protein count} & \tblval{$n \ge 30$ for scorer comparisons} \\
\addlinespace
\multicolumn{2}{@{}l}{\textbf{Hardware}} \\

\tblkey{Software stack} & \tblval{Python 3.12.12, PyTorch 2.9.0 (CUDA 12.8 build), NumPy 1.26.4, SciPy 1.17.0, healpy 1.19.0, mrcfile 1.5.4, scikit-image 0.26.0, scikit-learn 1.8.0} \\
\tblkey{Accelerator} & \tblval{NVIDIA A100 or A40, single GPU per run} \\
\bottomrule
\end{tabular}}
\endgroup
\end{table}


\end{document}